\documentclass[10pt,journal,cspaper,compsoc]{IEEEtran}
\usepackage{array} 
\usepackage{amsmath}
\usepackage{amssymb}
\usepackage{multirow}
\usepackage{caption}
\usepackage{hhline}
\usepackage{graphicx}
\usepackage{epstopdf} 
\usepackage{color}
\usepackage{graphicx}
\usepackage{subcaption}

\usepackage{times}
\usepackage{epsfig}
\usepackage{graphicx}
\usepackage{amsmath}
\usepackage{amssymb}
\usepackage{multirow}
\usepackage{hhline}
\usepackage{booktabs}
 
\usepackage[linesnumbered,ruled,vlined]{algorithm2e}

\usepackage{todonotes}
\usepackage{changepage}

\ifCLASSOPTIONcompsoc
\else
\fi

\def\etal{\emph{et al}~}

\ifCLASSINFOpdf
\else
\fi

\begin{document}
%
\title{Bayesian Joint Modelling for Object Localisation in Weakly Labelled Images}
\author{Zhiyuan Shi, Timothy M. Hospedales, Tao Xiang}
\IEEEcompsoctitleabstractindextext{%



\begin{abstract}
We address the problem of localisation of objects as bounding boxes in images and videos with weak labels. This weakly supervised object localisation problem has been tackled in the past using discriminative models where each object class is localised independently from other classes. In this paper, a  novel framework based on Bayesian joint topic modelling is proposed, which differs significantly from the existing ones in that: (1) All foreground object classes are modelled jointly in a single generative model that encodes multiple object co-existence so that ``explaining away'' inference can resolve ambiguity and lead to better learning and localisation. (2) Image backgrounds are shared across classes to better learn varying surroundings and ``push out'' objects of interest. (3) Our model can be learned with a mixture of weakly labelled and unlabelled data, allowing the large volume of unlabelled images on the Internet to be exploited for learning. Moreover, the Bayesian formulation enables the exploitation of various types of prior knowledge to compensate for the limited supervision offered by weakly labelled data, as well as Bayesian domain adaptation for transfer learning. Extensive experiments on the PASCAL VOC, ImageNet and YouTube-Object videos datasets demonstrate the effectiveness of our Bayesian joint model for weakly supervised object localisation.
\end{abstract}


}

\maketitle

\section{Introduction}

Object recognition is a challenging problem especially at a large scale  because of variabilities in object appearance, viewpoint, illumination and pose \cite{DengECCV2010,tim2011tpami,LempitskICCV2009}. Fully/strongly annotated data is thus typically required to learn a generalisable model for tasks such as object classification \cite{Nguyenweakly2011}, detection \cite{Felzenszwalb2012partbased,dollar08}, and segmentation \cite{LempitskICCV2009,Kuettel2012,Rubinstein13Unsupervised}. In fully annotated images, such as those in the PASCAL VOC object classification or detection challenges \cite{pascalvoc2007},  not only the presence of objects, but also their locations are labelled, typically in the form of bounding boxes. Such a strong manual annotation of objects is time-consuming and laborious. Consequently, although media data is increasingly available with the prevalence of sharing websites such as Flickr, the lack of annotated images, particularly strongly annotated ones, becomes the new barrier that prevents tasks such as object detection from scaling to thousands of classes \cite{Guillaumin_cvpr12}.   

\begin{figure}[ht]
\begin{center}
   \includegraphics[width=\linewidth]{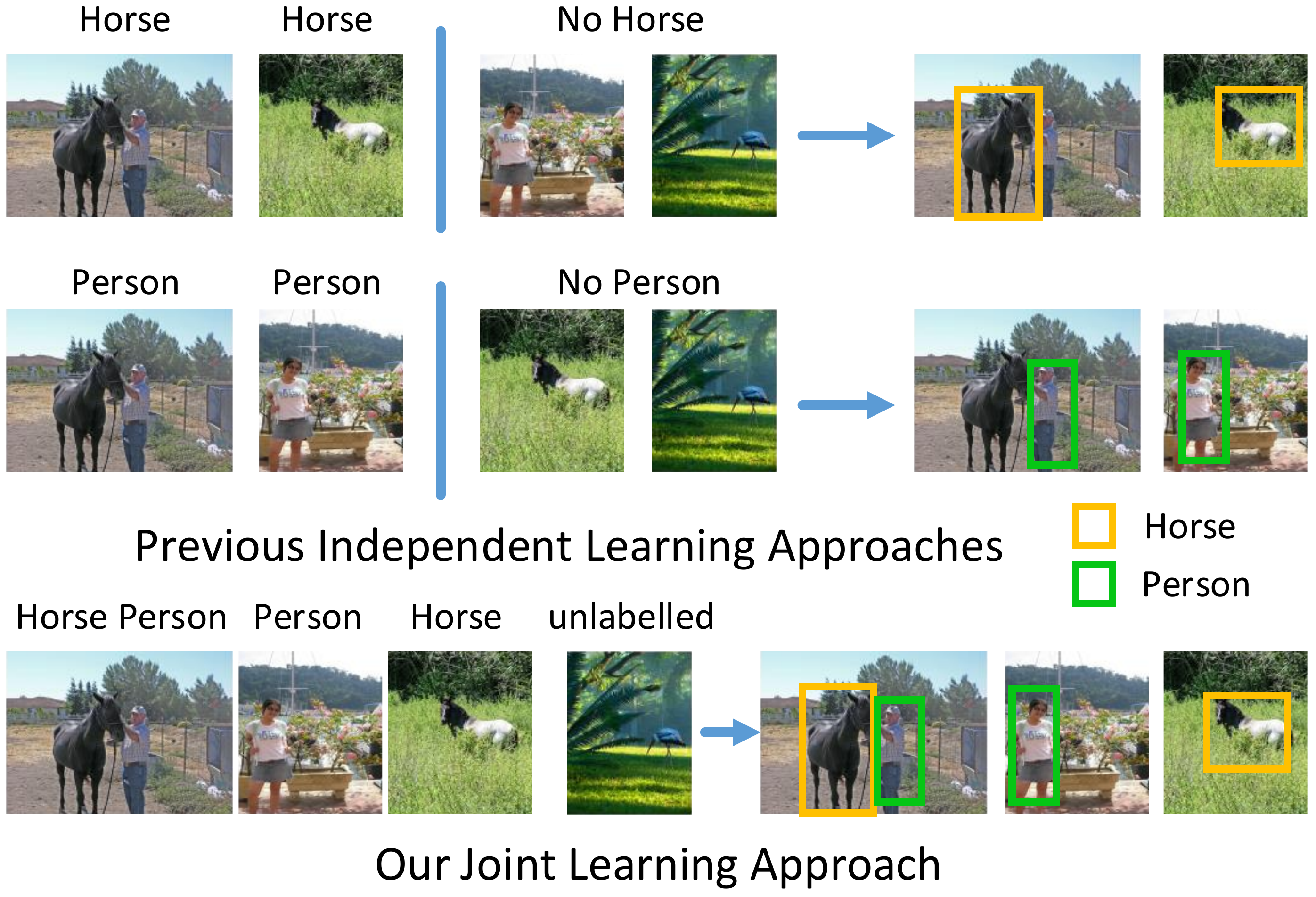}
\end{center}

   \caption{Different types of objects often co-exist in a single image. Our joint learning approach differs from previous approaches which localise each object class independently.}
\label{fig:concept}
\end{figure}

One approach to this challenge is weakly supervised object localisation (WSOL): simultaneously locating objects in images and learning their appearance using only weak labels indicating presence/absence of the objects of interest.
The WSOL problem has been tackled using various approaches \cite{Deselaers2012,Nguyenweakly2011,confeccvSivaRX12,Pandeyiccv2011,Guillaumin_cvpr12,Carolina2008eccv,TangCVPR14}. 
Most of them address the task as a weakly supervised learning problem, particularly as a multi-instance learning (MIL) problem, where images are bags, and potential object locations are instances. These methods are typically discriminative in nature and attempt to localise each class of objects independently from the other classes. However, localising objects of different classes independently has a number of limitations: 
(1) It fails to exploit the knowledge that different objects often co-exist within an image (see Fig.~\ref{fig:concept}). For instance, knowing that some images have both a horse and a person, in conjunction with a joint model for both classes -- the person can be ``explained away" to reduce ambiguity about the horse's appearance, and vice versa. Ignoring this  increases ambiguity for each class.
(2) Although  object classes vary in appearance, the background appearance is relevant to them all (e.g.~sky, tree, and grass are constant features of an image regardless of the foreground object classes). When different classes are modelled independently, the background must be re-learned repeatedly for each class, when it would be more statistically robust \cite{Salakhutdinov2011cvpr} to share this common knowledge. 

In this paper, a novel framework based on Bayesian latent topic models is proposed to overcome the mentioned limitations. In our framework, both multiple object
classes and background  types are modelled
jointly in a single generative model as latent topics, in order
to explicitly exploit their co-existence relationship (see Fig.~\ref{fig:concept}). As bag-of-words (BoW) models, conventional latent topic models
have no notion of localisation. We overcome this problem
by incorporating an explicit notion of object location. 


 Our generative model based framework has the following advantages over previous discriminative approaches:\\
\noindent  \textbf{Joint vs. independent modelling}\quad 
By jointly modelling different classes of objects and background, our model is able to exploit multiple object co-occurrence, so each object known to appear in an image can help disambiguate the location of the others by accounting for some of the pixels. This is illustrated by the left column of Fig.~\ref{fig:concept}, where modelling horse and person jointly helps the localisation of both objects since each single pixel can only be explained by one object, not both. 
Meanwhile, a single set of shared background topics are learned once for all object classes. This is due to the nature of a generative model -- every pixel in the image must be accounted for. Even though learning background appearance can further disambiguate the location of objects, this appears to be an extremely hard task given that no labels are provided regarding background (people tend to focus on the foreground when annotating an image). However, by learning them jointly with the foreground objects and using all training images available, this task can be fulfilled effectively by the proposed model.

\noindent  \textbf{Integration of prior knowledge}\quad 
Exploiting prior knowledge or top-down cues about appearance or geometry (e.g., position, size, aspect ratio) should be supported if available to offset the weak labels. 
 Our framework is able to incorporate, when available, prior knowledge about appearances of objects in a more systematic way as a Bayesian prior. 
Specifically, we exploit the prior intuition that objects are spatially compact relative to the background. We can also optionally exploit external human or internal data-driven prior about typical object size, location and appearance as a Bayesian prior. Going beyond within-class priors, we also show that cross-class appearance similarity can be exploited. For instance, the model can exploit the fact that ``bike'' is more similar to ``motorbike'' than ``aeroplane''. 

\noindent  \textbf{Bayesian domain adaptation}\quad 
A central challenge for building generally useful recognition models is providing the capability to adapt models trained on one domain or dataset to new domains or datasets \cite{eth_biwi_00905}. This is important because any given domain or dataset is intentionally or unintentionally biased \cite{Torralba_cvpr11}, so transferring models directly across domains generally performs poorly \cite{Torralba_cvpr11}. However, with appropriate adaptation, source and target domain data can be combined to out-perform target domain data alone \cite{eth_biwi_00905}. We can leverage our model's Bayesian formulation to provide domain adaptation in a WSOL context. 



\noindent  \textbf{Semi-supervised learning}\quad
Since there are effectively unlimited quantity of unlabelled data available on the Internet (compared to limited quantity of manually annotated data), a valuable capability is to exploit this existing unlabelled data in conjunction with limited weakly labelled data to improve learning. 
As a generative model, our framework is naturally suited for semi-supervised learning (SSL). Unlabelled data are included and the label variables for these instances left unclamped (i.e.~no supervision is enforced). Importantly, unlike conventional SSL approaches \cite{zhu2007sslsurvey}, our model does not require that all the unlabelled data are instances of known classes, making it more applicable to realistic SSL applications.



\section{Related Work}
\label{relatedwork}
\noindent  \textbf{Weakly supervised object localisation}\quad Weakly supervised learning (WSL) has attracted increasing attention as the volume of data which we are interested in learning from grows much faster than available annotations. Weakly supervised object localisation (WSOL) is of particular interest \cite{Deselaers2012,Sivaiccv2011,confeccvSivaRX12,TangCVPR14,Pandeyiccv2011,zhiyuan12,Nguyenweakly2011,Crandalleccv06,TangCVPR14,TangECCV14}, due to the onerous demands of annotating object location information. Many studies \cite{Nguyenweakly2011,Deselaers2012} have approached this task as a multi-instance learning  \cite{Maron98aframework,Andrews03supportvector} problem. However, only relatively recently have localisation models capable of learning from challenging data such as the PASCAL VOC 2007 dataset been proposed \cite{Deselaers2012,Sivaiccv2011,confeccvSivaRX12,Pandeyiccv2011,zhiyuan12}. Such data is especially challenging because objects may occupy only a small proportion of an image, and multiple objects may occur in each image: corresponding to a multi-instance multi-label problem \cite{nguyen2010svm_miml}. Three types of cues are exploited in existing WSL object localisation approaches: (1) \textit{saliency} -- a region containing an object should look different from the majority of (background) regions. The object saliency model in  \cite{Alexe_TPAMI_2012} is  widely used in most recent work \cite{Deselaers2012,confeccvSivaRX12,Guillaumin_cvpr12,Sivaiccv2011,eth_biwi_00905} as a preprocessing step to propose a set of candidate object locations so that the subsequent computation is reduced to a tractable level, (2) \textit{intra-class} -- a region containing an object should look similar to the regions containing the same class of objects in other  images \cite{Sivaiccv2011}, and (3) \textit{inter-class} -- the region should look dissimilar to any regions that are known to not contain the object of interest \cite{Deselaers2012,confeccvSivaRX12,Pandeyiccv2011}. One of the first studies to combine the three cues for WSOL was \cite{Deselaers2012} which employed a conditional random field (CRF) and generic prior object knowledge learned from a fully annotated dataset. Later, \cite{Pandeyiccv2011} presented a solution exploiting latent SVMs. Recent studies have explicitly examined the role of intra- and inter-class cues  \cite{Sivaiccv2011,confeccvSivaRX12}, as well as transfer learning \cite{zhiyuan12,Guillaumin_cvpr12}, for this task. Similar to the above approaches for weakly labelled images, \cite{Tang_NIPS2012,eth_biwi_00905} proposed video based frameworks to deal with motion segmented tubes instead of bounding-boxes. In contrast to these studies, which are all based on discriminative models, we introduce a generative topic model based approach that exploits all three cues, as well as joint multi-label, semi-supervised and cross-domain adaptive learning. 

\noindent  \textbf{Exploiting prior knowledge} \quad Prior knowledge has been exploited in existing WSOL works \cite{Deselaers2012,confeccvSivaRX12,Pandeyiccv2011}. Recognition or detection priors can be broadly broken down into appearance and geometry (location, size, aspect) cues, and can be provided manually, or estimated from data. The most common use has been crude: to generate candidate object locations based on a pre-trained model for generic objectness \cite{Alexewhatisobject}, i.e.~the previously mentioned saliency cue. This reduces the search space for discriminative models. Beyond this, geometry priors have also been estimated during learning \cite{Deselaers2012}. 
We can not only exploit such simple appearance and geometry cues as model priors, but also go beyond to exploit a richer object hierarchy, which has been widely exploited in classification \cite{danielcvpr2012,zweig07_iccv,Rohrbach2011cvpr,Salakhutdinov2011cvpr} and to a less extent detection \cite{Guillaumin_cvpr12,Kuettel2012}.
More specifically, we leverage WordNet, a large lexical database based on linguistics \cite{Pedersen2004}. WordNet provides a measure of prior appearance similarity/correlation between classes, and we use this prior to regularise appearance learning. Such cross-class appearance correlation information is harder to use in previous WSOL approaches because different classes are trained separately. 
Interestingly, our model uniquely shows positive results for WordNet-based appearance correlation (see Sec.~\ref{soa}), in contrast to some recent studies \cite{Rohrbach2011cvpr,Salakhutdinov2011cvpr} that found no or limited benefit from exploiting WordNet based cross-class appearance correlation for recognition. Compared to the classification task, this inter-class correlation information is more valuable for WSOL because the task is more ambiguous. Specifically, the interdependent localisation and appearance learning aspects of the task adds an extra layer of ambiguity -- the model might be able to learn the appearance if
it knew the location, but it will never find the location without knowing appearance.
Our work is related to \cite{Guillaumin_cvpr12} where hierarchical cross-class appearance similarity is used to help weakly supervised object localisation in ImageNet by transfer learning. However, a source dataset of fully annotated images are required in their work, whilst our model exploits the correlation directly for the target data which is only weakly labelled. 

\noindent  \textbf{Cross domain/dataset learning}\quad
Domain adaptation \cite{Cao2010cvpr} methods aim to exploit prior knowledge from a source domain/dataset to improve the performance and/or reduce the amount of annotation required in a target domain/dataset (see \cite{pan2009transfer_survey} for a review). Many conventional approaches are based on SVMs for which the target domain can be considered a perturbed version of the source domain, and thus learning proceeds in the target domain by regularising it toward the source \cite{yang2007crossDomainConcept}. More recently, transductive SVM \cite{BergamoTorresani10}, Multiple Kernel Learning (MKL) \cite{Luo2011iccv}, and instance constraints \cite{Donahue_CVPR2013} have been exploited. In contrast to these discriminative approaches, we exploit a simple and efficient Bayesian adaptation approach similar in spirit to \cite{Cao2010cvpr,Dai2007aaai}. Posterior parameters from the source domain are transferred as priors for the target, which are then adapted based on observed target domain data via Bayesian learning. Going beyond simple within-modality dataset bias, recent studies   \cite{eth_biwi_00905,Tang_NIPS2012} have adapted object detectors from video to image or reverse. We show that our approach can achieve the image-video domain transfer within a single framework.

\noindent  \textbf{Exploiting unlabelled data}\quad Semi-supervised learning \cite{zhu2007sslsurvey} methods aim to  reduce labelling requirements and/or improve results compared to only using labelled data. Most existing SSL approaches assume  a training set with a mix of fully labelled and weak or unlabelled \cite{blaschko10simultaneous,Guillaumin_cvpr12} data, while we use weak and unlabelled data alone. The existing (discriminative) line of work focusing on WSOL \cite{Deselaers2012,Pandeyiccv2011,Tang_CVPR2013,Glenn2012eccv} has not generally exploited unlabelled data, and cannot straightforwardly do so.


\noindent  \textbf{Topic models for image understanding}\quad Latent topic models (LTMs) were originally developed for unsupervised text analysis \cite{BleiLDA2003}, and have been successfully adapted to both unsupervised \cite{Philbinijcv2010,Sivic05b} and supervised image understanding problems \cite{feifei2006one_shot,CaoFei-Fei2007,LiSocherFeiFei2009,wangbleifeifei08,Rasiwasia_TPAMI_2013}. Most studies have addressed the simpler tasks of classification \cite{wangbleifeifei08,Rasiwasia_TPAMI_2013} and annotation \cite{wangbleifeifei08,blei2003annotated_model}.
Our model differs from the existing ones in two main aspects: (i) Conventional topic models have no explicit notion of the spatial location and extent of an object in an image. This is addressed in our model by modelling the spatial distribution of each topic. Note that some topic model based methods \cite{LiSocherFeiFei2009,CaoFei-Fei2007} can also be applied to object localisation. However, the spatial location is obtained from a pre-segmentation step rather than being explicitly modelled.  (ii) The other difference is more subtle -- existing supervised topic models such as CorrLDA \cite{blei2003annotated_model}, SLDA \cite{wangbleifeifei08} and derivatives \cite{LiSocherFeiFei2009} only weakly influence the learned topics. This is because the objective is the sum of visual words and label likelihoods, and  visual words vastly outnumber annotations, thus dominating the result \cite{Rasiwasia_TPAMI_2013}. The limitation is serious for WSOL as the labels are already weak and they must be used to their full strength. In this work, a learning algorithm with topic constraints similarly to \cite{tim2011tpami} is formulated to provide stronger supervision which is demonstrated to be much more effective than the conventional supervised topic models in our experiments (see supplementary material). With these limitations addressed, we can exploit the potential of a generative model for domain adaptation, joint-learning of multiple objects and semi-supervised learning.

\noindent \textbf{Other joint learning approaches}\quad An approach similar in spirit to ours in the sense of jointly learning a model for all classes is that of Cabral \etal \cite{CabralDCB11}. This study formulates multi-label image classification as a matrix completion problem, which is also related to our factoring images into a mixture of topics. However we add two key components of (i) a stronger notion of the spatial location and extent of each object, and (ii) the ability to encode human knowledge or transferred knowledge through a Bayesian prior. As a result, we are able to address more challenging data than \cite{CabralDCB11} such PASCAL VOC.
Multi-instance multi-label (MIML) \cite{nguyen2010svm_miml} approaches provide a mechanism to jointly learn a model for all classes \cite{Zhou07multi-instancemultilabel,zha2008mlmi}. However, because these methods must search for a discrete space (of positive instance subsets), their optimisation problem is harder than the smooth probabilistic optimisation here.
Finally, while more elaborate joint generative learning methods \cite{sudderth2008tdp_visual,LiSocherFeiFei2009} exist, they are more complicated than necessary for WSOL and do not scale to the size of data required here.

\noindent  \textbf{Feature fusion}\quad Combining multiple complementary cues has been shown to improve classification performance in object recognition \cite{GehlerN09,Gehler09_let,Orabona10,Luo2011iccv}. Two simple feature fusion methods have been widely used in existing work: early fusion which combines low-level features \cite{Siva_2013_CVPR} early (feature concatenation) and late (score level) fusion \cite{Deselaers2012,Sivaiccv2011}. Multiple kernel learning (MKL) approaches have attracted  attention as a principled mid-level approach to combining features \cite{Orabona10,Gehler09_let}. Similarly to MKL, our framework provides a principled and jointly-learned mid-level probabilistic fusion via its generative process. 

\noindent  \textbf{Contributions}\quad In summary, this paper makes the following contributions: (1) We propose the novel concept of joint modelling of all object classes and background for weakly supervised object localisation. (2) We formulate a novel Bayesian topic model suitable for object localisation, which can use  various types of prior knowledge including an  inter-category appearance similarity prior.  (3) Our Bayesian prior enables the model to easily borrow  available domain knowledge from existing auxiliary datasets and adapt it to a target domain. (4) We further exploiting unlabelled data for improving weakly supervised object localisation. (5) Extensive experiments on the PASCAL VOC 2007 \cite{pascalvoc2007} and ImageNet \cite{imagenet_cvpr09} show that our model surpasses existing competitors and achieves state-of-the-art performance. A preliminary version of our work was described in \cite{shi_iccv_2013}.


\section{Joint Topic  Model for Objects and Background}\label{sec:PGM}

 In this section, we introduce our new latent topic model (LTM) \cite{BleiLDA2003} approach to the weakly-supervised object localisation task. Applied to images, conventional LTMs factor images into combinations of latent topics \cite{Philbinijcv2010,Sivic05b}. Without supervision, these topics may or may not correspond to anything of semantic relevance to humans. To address the WSOL task, we need to learn what is unique to all images sharing a particular label (object class), while explaining away both the pixels corresponding to other annotated objects as well as  other shared visual aspects (background) which are irrelevant to the annotations of interest. We will achieve this in a LTM framework by applying weak supervision to partially constrain the available topics for each image. This constraint is enforced by label/topic clamping to ensure that each foreground topic corresponds to an object class of interest.

\begin{figure}[t]
\begin{center}
   \includegraphics[width=1.0\columnwidth]{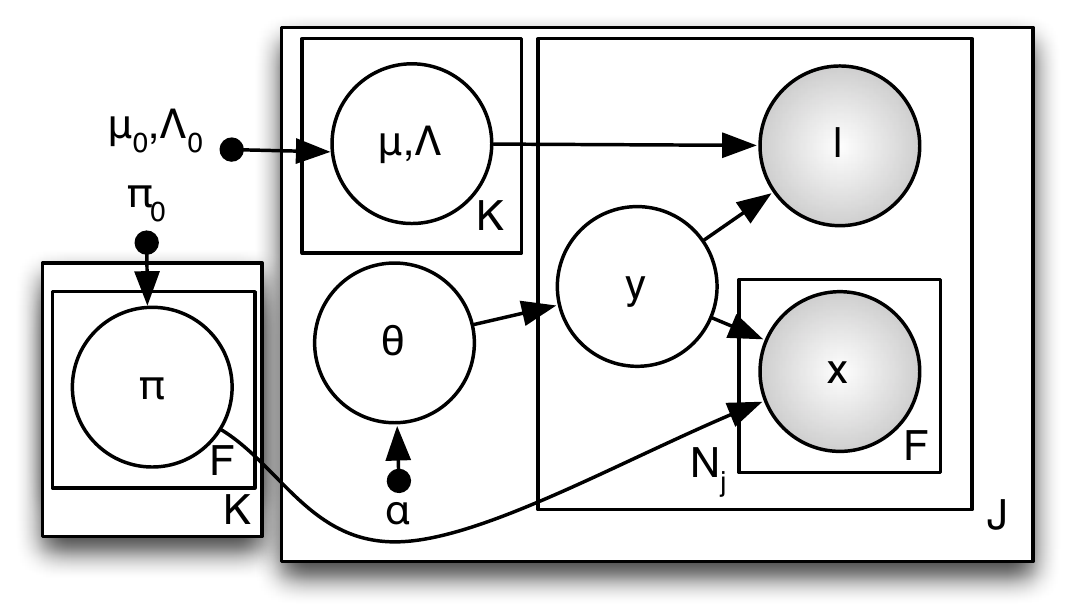}
\end{center}
   \caption{Graphical model for our WSOL joint topic model. Shaded nodes are observed.}
\label{fig:GM}
\end{figure}

More specifically, to address the WSOL task, we will factor images into unique combinations
of $K$ shared topics. If there are $C$ classes of objects to
be localised, $K^{fg}=C$ of these will represent the (foreground) classes,
and $K^{bg}=K-K^{fg}$ topics will model background
data to be explained away. Each topic thus corresponds to one object class or one type of background. Let $T^{fg}$ and $T^{bg}$ index foreground
and background topics respectively. An image is represented using a Bag-of-Words (BoW)  representation for each of $f=1\dots F$ different types of features (see Sec.~\ref{datasets} for the specific  appearance features used). After learning, each latent topic will encode both a distribution
over the $V_f$ sized appearance vocabulary of each feature $f$ and also over the spatial
location of these words within each image. Formally, given a set of $J$ training images, each labelled with any number of the $C$ foreground classes, and represented as bags of words $\mathbf{x}_{jf}$, the generative
process of our model (Fig.~\ref{fig:GM})  is as follows (notation is summarised in Table \ref{trainandtest} for convenience):
\\~\\
\noindent For each topic $k\in1\dots K$:
\begin{enumerate}
   \item  For each feature representation $f\in1\dots F$:
   \begin{enumerate}   
     \item Draw an appearance distribution $\boldsymbol{\pi}_{kf}\sim\mbox{Dir}(\boldsymbol{\pi}_{kf}^{0})$ following the Dirichlet distribution
     \end{enumerate}  
\end{enumerate}
For each image $j\in1\dots J$:
\begin{enumerate}
\item Draw foreground and background topic distribution $\boldsymbol{\theta}_{j}\sim\mbox{Dir}(\boldsymbol{\alpha}_{j})$,
$\boldsymbol{\alpha}_{j}=[\boldsymbol{\alpha}_{j}^{fg},\boldsymbol{\alpha}_{j}^{bg}]$ where the Dirichlet distribution parameter $\alpha_{j}$ reflects prior knowledge of the presence of each object class or background in the image $j$. Both $\boldsymbol{\theta}_{j}$ and $\boldsymbol{\alpha}_{j}$ are $K$ dimensional.
\item For each foreground topic $k\in T^{fg}$ draw a location distribution:
$\{\boldsymbol{\mu}_{kj},\Lambda_{kj}\}\sim\mathcal{NW}(\boldsymbol{\mu}_{k}^{0},\Lambda_{k}^{0},\beta_{k}^{0},\nu_{k}^{0})$
\item For each observation (visual word) $i\in1\dots N_j$:

\begin{enumerate}
\item Draw topic $y_{ij}\sim\mbox{Multi}(\boldsymbol{\theta}_{j})$
\item Draw a location: \\$\mathbf{l}_{ij}\sim\mathcal{N}(\boldsymbol{\mu}_{y_{ij}j},\Lambda_{y_{ij}j}^{-1})$
if $y_{ij}\in T^{fg}$ or\\ $\mathbf{l}_{ij}\sim Uniform$ if $y_{ij}\in T^{bg}$


\item For each feature representation $f\in1\dots F$:

\begin{enumerate}

\item Draw visual word $x_{ijf}\sim\mbox{Multi}(\boldsymbol{\pi}_{y_{ij}f})$
\end{enumerate}

\end{enumerate}
\end{enumerate}
\noindent where Multi, Dir, $\mathcal{N}$, $\mathcal{NW}$ and $Uniform$ respectively indicate Multinomial, Dirichlet, Normal, Normal-Wishart and uniform distributions with the specified parameters. These prior distributions are chosen mainly because they are conjugate to the word, topic and location distributions, and hence enable efficient inference. For the visual word spatial location,  the foreground and background distributions are of different forms -- normal for foreground  and uniform for background. This is to reflect the intuition that foreground objects tend to be compact and background much less so. 
The joint distribution of all observed $O=\{\mathbf{x}_{jf},\mathbf{l}_{j}\}_{j,f=1}^{J,F}$
and latent $H=\{\{\boldsymbol{\pi}_{kf}\}_{k,f=1}^{K,F},\{\mathbf{y}_{j},\boldsymbol{\mu}_{kj},\Lambda_{kj},\boldsymbol{\theta}_{j}\}_{k,j=1}^{K,J}\}$
variables given parameters $\Pi=\{\{\boldsymbol{\pi}_{kf}^{0}\}_{k,f=1}^{K,F},\{\boldsymbol{\mu}_{k}^{0},\Lambda_{k}^{0},\beta_{k}^{0},\nu_{k}^{0}\}^K_{k=1},\{\boldsymbol{\alpha}_{j}\}_{j=1}^{J}\}$
in our model is therefore:
\setlength\arraycolsep{1.5pt}
\begin{eqnarray}
p(O,H|\Pi)&=&\prod_{k}^{K}\prod_{f}^{F}p(\boldsymbol{\pi}_{kf}|\boldsymbol{\pi}_{kf}^{0})\\
&& \hspace{-2.1cm} \cdot \prod_{j}^{J}p(\boldsymbol{\theta}_{j}|\boldsymbol{\alpha}_{j})\left[\vphantom{\prod_{i}^{N_{i}}}\prod_{k}^{K}p(\boldsymbol{\mu}_{jk},\Lambda_{jk}|\boldsymbol{\mu}_{k}^{0},\Lambda_{k}^{0},\beta_{k}^{0},\nu_{k}^{0}) \right.\nonumber      \\
 && \left.\hspace{-2.1cm}\left(\prod_{i}^{N_j}p(\mathbf{l}_{ij}|\boldsymbol{\mu}_{jk},\Lambda^{-1}_{jk})\prod_{f}^{F}p(x_{ijf}|y_{ij},\boldsymbol{\pi}_{y_{ij}f})p(y_{ij}|\theta_{j})\right)\right].
 \label{eq:Joint}
\end{eqnarray}


\begin{table}[t]
\setlength{\heavyrulewidth}{0.12em}
\centering
\begin{tabular}{ll}
\toprule
$x_{ijf}=1...V_f$  & Visual word $i$ in image $j$ for feature $f$ \\
\midrule
$\mathbf{l}_{ij}$ & Location of visual word $i$ in image $j$\\
\midrule
$y_{ijk} =1\dots K$  & Topic (object) for explaining visual word $x_{ijf}$  \\
\midrule
$\boldsymbol{\alpha}_j$ & Annotation / topic prior for image $j$\\
\midrule
$\boldsymbol{\theta}_{j}$ & Dirichlet topic proportion in image j\\
\midrule
$\boldsymbol{\pi}_{kf}^{0}$ & Appearance prior for topic/class $k$ in feature $f$\\
\midrule
 $\boldsymbol{\pi}_{kf}$ & Dirichlet appearance for topic/class $k$ in feature $f$\\
\midrule
 $\boldsymbol{\mu}_k^{0}, \Lambda_k^{0}$ & $\mathcal{NW}$ Location prior for class $k$\\
\midrule 
$\boldsymbol{\mu}_{kj}, \Lambda^{-1}_{kj}$ & Gaussian location of object class $k$ in image $j$\\
\bottomrule
\end{tabular}
\caption{Summary of model variables and parameters}
\label{trainandtest}
\end{table}

\section{Model learning}\label{sec:learning}

\noindent \textbf{Inference via variational message passing}\quad 
Learning our model involves inferring the following quantities: the
appearance of each object class for each feature type, $\boldsymbol{\pi}_{kf},k\in T^{fg}$ and each background
type, $\boldsymbol{\pi}_{kf},k\in T^{bg}$ for each feature type $f$; the word-topic distribution (soft segmentation) of each image $\mathbf{z}_{j}$,
the proportion of visual words (related to the proportion of pixels) in each image corresponding to each
class or background $\boldsymbol{\theta}_{j}$, and the location of each object
$\boldsymbol{\mu}_{jk},\Lambda_{jk}$ in each image (mean and covariance of a Gaussian). To learn the model and localise all the weakly
annotated objects, we wish to infer the posterior $p(H|O,\Pi)=p(\{\mathbf{y}_{j},\boldsymbol{\mu}_{jk},\Lambda_{jk},\boldsymbol{\theta}_{j}\}_{k,j}^{K,J},\{\boldsymbol{\pi}_{kf}\}_{k,f}^{K,F}|\{\mathbf{x}_{jf},\mathbf{l}_{j}\}_{j=1,f=1}^{J,F},\Pi)$.
This is intractable to solve directly; however a variational message passing
(VMP) \cite{winn2004vmp} strategy can be used to obtain a factored
approximation $q(H|O,\Pi)$ to the posterior:
\begin{eqnarray}
q(H|O,\Pi) & =\nonumber \\
 &  & \hspace{-1.5cm}\prod_{k,f}q(\boldsymbol{\pi}_{kf})\prod_{j}q(\boldsymbol{\theta}_{j})q(\boldsymbol{\mu}_{jk},\Lambda_{jk})\prod_{i}q(y_{ij}).\label{eq:varApprox}
\end{eqnarray}
\noindent Under this approximation a VMP solution is obtained by deriving integrals of the
form $\ln q(\mathbf{h})=E_{H\backslash\mathbf{h}}\left[\ln p(H,O)\right]+K$
for each group of hidden variables $\mathbf{h}$, thus obtaining the following 
updates for the sufficient statistics (indicated by tilde) of each variable:

\begin{eqnarray}
\tilde{\theta}_{jk} & = & \alpha_{jk}+\sum_{i} \tilde{y}_{ijk},\label{eq:theta}  \\
\tilde{y}_{ijk} & \propto & \int_{\boldsymbol{\mu}_{jk},\Lambda_{jk}}\mathcal{N}(\mathbf{l}_{ij}|\boldsymbol{\mu}_{jk},\Lambda_{jk}^{-1})q(\boldsymbol{\mu}_{jk},\Lambda_{jk}) \nonumber \\
 &  & \hspace{-1cm}\cdot\prod_{f}^{F}\exp\left(\Psi(\tilde{\pi}_{x_{ijf}y_{ij}f})-\Psi(\sum_{v}\tilde{\pi}_{vy_{ij}f}) \right)   \nonumber \\
 &  & \hspace{-1cm} \cdot\exp\left(\Psi(\tilde{\theta}_{jy_{ijk}})\right), \label{eq:y} \\
\tilde{\pi}_{vkf} & = & \pi^0_{vkf}+\sum_{ij}\mathbf{I}(x_{ijf}=v) \tilde{y}_{ijk}, \label{eq:varUpdates}
\end{eqnarray}


\noindent where $\Psi$ is the digamma function, $v=1\dots V_f$ ranges over the BoW appearance vocabulary, $\mathbf{I}$ is the indicator function which returns 1 if its argument is true, and
the integral in the second line returns the student-t distribution over $\mathbf{l}_{ij}$, $\mathcal{S}(\mathbf{l}_{ij}|\tilde{\boldsymbol{\mu}}_{jk},\tilde{\Lambda}_{jk}^{-1}P,\tilde{\beta}_{jk},\tilde{\nu}_{jk})$. Within each image $j$, standard updates \cite{bishop2006prml} apply for the sufficient statistics $\{\tilde{\boldsymbol{\mu}}_{jk},\tilde{\Lambda}_{jk},\tilde{\beta}_{jk},\tilde{\nu}_{jk}\}$ of the Normal-Wishart parameter posterior $q(\boldsymbol{\mu}_{jk},\Lambda_{jk})$. The update in Eq.~(\ref{eq:y}) (estimating the object explaining each pixel) is the most non-standard for LTMs; this is because it contains a top-down contribution (the third term), and two bottom-up contributions from the location and appearance (the first and second terms respectively). The model is learned by iterating the updates of Eqs.~(\ref{eq:theta})-(\ref{eq:varUpdates}) for all images $j$, words $i$, topics $k$ and vocabulary $v$.

\noindent \textbf{Supervision via label-topic constraints}\quad 
In conventional topic models, the $\alpha$ parameter encodes the expected proportion of words for each topic. In our weakly supervised topic model, we use $\alpha$ to encode the supervision from weak labels. In particular, we set $\alpha_{j}^{fg}$
as a binary vector with $\alpha_{jk}^{fg}=1$ if class $k$ is present
in image $j$ and $\alpha_{jk}^{fg}=0$ otherwise. $\alpha^{bg}$
is always set to $1$ to reflect the fact that background of different types can be shared across different images. That is, the foreground topics are clamped with the weak labels indicating the presence/absence of foreground object classes in each image, whilst all background types are assumed to be present a priori.  With these partial constraints, iterating the
updates in Eqs.~(\ref{eq:theta})-(\ref{eq:varUpdates}) has the effect of factoring images
into combinations of latent topics, where $K^{bg}$ background
topics are always available to explain away backgrounds, and
$K^{fg}$ foreground topics are only available to images with
annotated classes. Note that this set-up assumes a 1:1 correspondence between object classes and topics. More topics can trivially be assigned to each object class (1:N correspondence), which has the effect of modelling multi-modality in object appearance, for a linear increase in computational cost.

\noindent \textbf{Probabilistic feature fusion}\quad 
We combine multiple features probabilistically in our model. A single topic distribution ($y$) is estimated given  different low-level features ($f$) in  Eq.~(\ref{eq:y}). Our fusion keeps the original low-level feature representations rather than increasing ambiguity by concatenating them before they provide complementary information about the location (early fusion). The shared topic ($y$) and Gaussian location distribution ($\boldsymbol{\mu},\Lambda^{-1}$) correlate the multiple features, which avoids  domination by a single one. The appearance model in each modality is updated based on the consensus estimate of location; it thus learns a good appearance in each view even if the particular category is hard to detect in that view (as a result could drift if used alone). Its advantage over early (feature concatenation) or late (score level) fusion is demonstrated experimentally in Sec.~1 of the supplementary material.

\section{Object Localisation\label{sub:Object-Localisation}}

After learning, we extract the location of the objects in each image from the model, which can then be used to learn an object detector. Depending on whether the images are treated as individual images or consecutive video frames, our localisation method differs slightly.

\noindent \textbf{Individual images}\quad There are two possible strategies to localise objects in individual images, which we will compare later in Sec.~\ref{experiments}. In the first strategy (\textit{Our-Gaussian}), a bounding box
for class $k$ in image $j$ can be obtained directly from the Gaussian mean of the 
parameter posterior $q(\boldsymbol{\mu}_{jk},\Lambda_{jk})$,
via aligning a bounding box to the two standard deviation ellipse.  This has the advantage of being
clean and highly efficient. However, since there
is only one Gaussian per class (which will grow to cover all instances
of the class in an image), this is not ideal for images with more
than one object per class. In the second strategy (\textit{Our-Sampling}) we draw a heat-map for class $k$ by projecting $q(y_{ijk})$ (Eq.~(\ref{eq:y})) 
back onto the image plane,
using the grid coordinates where visual words are computed. This heat-map is analogous to those
produced by many other approaches such as Hough transforms \cite{houghforest2012}.
Thereafter, any strategy for heat-map based localisation may be used.
We choose the non-maximum suppression (NMS) strategy of \cite{Felzenszwalb2012partbased}.

\noindent \textbf{Video frames}\quad The above two strategies are directly applicable to video data if we treat each frame as an individual image. However, the temporal information of objects is useful in continuous videos to smooth  the noise of individual frames. To this end, we apply a simple state space model for video segments to post-process object locations, smoothing them in time. Two diagonal points are sufficient to encode object location (bounding-box), and these are estimated from  $q(\boldsymbol{\mu}_{jk},\Lambda_{jk})$ above at every frame/time $t$ as $c_t$. Assuming a four-dimensional state latent state vector $\mathbf{z}_{t}^{T}=(z_{xt} \ z_{yt} \ \dot{z}_{xt} \ \dot{z}_{yt})$,  denoting the (hidden) true coordinate of an object of interest (two diagonal corners of the bounding box). A Kalman smoother is then adopted to smooth the observation noise $\sigma_t$  in the system:

\begin{equation}
\mathbf{z}_{t} = \mathbf{A}\mathbf{z}_{t-1}+\epsilon_t,\\~\\
\mathbf{c}_t = \mathbf{O}\mathbf{z}_t + \sigma_t,
\end{equation}

\noindent where $\mathbf{A}$ is the temporal transition between true locations $\mathbf{z}$ in video, and $\mathbf{O}$ is the observation function for each frame.


\section{Bayesian Priors}

An important capability of our Bayesian approach is that top-down cues from human expertise, or estimated from data can be encoded. Various types of human knowledge about objects and their relationships with background are encoded in our model. As discussed earlier, prior cues can potentially cover appearance and geometry information.

\noindent \textbf{Encoding geometry prior}\quad For geometry, we already model the most general intuition that objects are compact relative to background by assigning them Gaussian and uniform distributions respectively (Sec.~\ref{sec:PGM}). Beyond this, prior knowledge about typical image location and size of each class can be included via prior parameters $\boldsymbol{\mu}^0_{k},\Lambda^0_k$, however we found this did not actually noticeably improve results in our experiments so we did not exploit it. This makes sense, because in challenging datasets like PASCAL VOC, objects appear in highly variable scales and locations, so there is little regularity to learn.

\noindent \textbf{Encoding appearance prior}\quad If prior information is available about object category appearance, it can be included by setting $\boldsymbol{\pi}^0_{kf}$. (We will exploit this later for cross-domain adaptation in Sec.~\ref{sec:BDA}). For within-domain learning, we can obtain an initial data-driven estimate of object appearance to use as a prior by exploiting the observation that, when aggregated across all images, the background is more dominant than any single object class in terms of size (hence the amount of visual words). Exploiting this intuition, for each object class $k$, we set the appearance prior $\boldsymbol{\pi}^0_{kf}$ as:
\begin{equation}
\boldsymbol{\pi}^0_{kf} = \left|\frac{1}{C}\sum_{j,c_j=k} h(\mathbf{x}_{jf})-\frac{1}{J}\sum_j h(\mathbf{x}_{jf}) \right|_++ \epsilon,\label{eq:appPrior}
\end{equation}
 where $h(\cdot)$ indicates histogram and $\epsilon$ is a small constant. That is, set the appearance prior for each class to the mean of those images containing the object class minus the average over all images. This results in a prior which reflects what is consistently unique about each particular class. This is related to the notion of saliency, not within an image, but across all images. Saliency has been exploited in previous MIL based approaches to generate the instances/candidate object locations \cite{Deselaers2012,Sivaiccv2011,confeccvSivaRX12,Pandeyiccv2011,zhiyuan12}. However, in our model it is cleanly integrated as a prior.

\noindent \textbf{Encoding appearance similarity prior}\quad
Going beyond the direct unary appearance prior discussed above, we next consider exploiting the notion of prior \emph{inter-class appearance similarity}, rather than prior appearance per-se. The prior similarity between each object category can be estimated by computing inter-category category distance based on WordNet structure \cite{Pedersen2004}. We compute a similarity matrix $\mathcal{M}$ where elements $\mathcal{M}_{m,n}$ indicates the relatedness between class $m$ and $n$. The similarity matrix is then used to define how much appearance information from class $m$ contributes to class $n$ a priori. 

We exploit this matrix by introducing an M-step into our learning algorithm  (Eqs.~(\ref{eq:theta})-(\ref{eq:varUpdates})). Previously the appearance prior $\boldsymbol{\pi}_{kf}^0$ was considered fixed (e.g., from Eq.~(\ref{eq:appPrior})). As with any parameter learning in the presence of latent variables, $\boldsymbol{\pi}_{kf}^0$ could potentially be optimised by a maximum-likelihood M-step interleaved with E-step latent variable inference. However, rather than the conventional approach of optimising $\boldsymbol{\pi}_{kf}^0$ \emph{solely} given the data of class $k$, we define an update that exploits cross-class similarity by updating  $\boldsymbol{\pi}_{kf}^0$  using \emph{all} the data, but weighted by its similarity to the target class $k$.

 Denoting $\hat{{\pi}}^0_{vkf}$ as the new appearance  prior to be learned, we introduce a new regularised M-step to learn $\hat{{\pi}}^0_{vkf}$. Specifically, the update for each class $k\in T^{fg}$ is as follows:
\begin{equation}
\hat{\pi}^0_{vkf}=\hspace{-0.3cm}\underbrace{\pi_{vkf}^{0}}_\text{fixed data driven prior}\hspace{-0.1cm}+\underbrace{\sum_{ij}\sum_{k' \in T^{fg}} \mathcal{M}_{k,k'} \cdot \mathbf{I}(x_{ijf}=v) \tilde{y}_{ijk'}}_\text{inter-class similarity prior}
\end{equation}
The first term $\pi_{vkf}^{0}$ is the original unary prior from Eq.~(\ref{eq:appPrior}). The second term is a data-driven update given the results of the E-step ($\tilde{y}$, Eqs.~(\ref{eq:theta})-(\ref{eq:varUpdates})). It includes a contribution from all images of all classes $k'$, weighted by the similarity of $k'$ to the target class $k$ -- given by $\mathcal{M}_{k,k'}$. The updated $\hat{\boldsymbol{\pi}}^0_{kf}$ then replaces $\boldsymbol{\pi}^0_{kf}$ in  Eq.~(\ref{eq:varUpdates}) of the E-step.

%

\section{Learning from additional data}

In this section, we discuss learning from additional data beyond the data for the WSOL task. This includes partially relevant data from other domains or datasets, and any additional but un-annotated data from the same domain.

\subsection{Bayesian Domain Adaptation}\label{sec:BDA}

Across different datasets or domains (such as images and video), the appearance of each object category will exhibit similarity, but vary sufficiently that directly using an appearance model learned in a source domain $s$ for inference in a target domain $t$ will perform poorly \cite{Torralba_cvpr11}. In our case this would correspond to directly applying a learned source appearance model $\pi^s_k$ to a new target domain $t$, $\boldsymbol{\pi}^t_k:=\boldsymbol{\pi}^s_k$. However, one hopes to be able to exploit similarities between the domains to learn a better model than using only the target domain alone \cite{yang2007crossDomainConcept,BergamoTorresani10,Tang_NIPS2012,Dai2007aaai}. In our case, the Bayesian (Multinomial-Dirichlet conjugate) form of our model is able to  achieve this for WSOL by simply learning $\boldsymbol{\pi}^s_k$ for a source domain $s$ (Eq.~(\ref{eq:varUpdates})), and applying it as the prior ${\boldsymbol{\pi}_k^0}^t:=\boldsymbol{\pi}^s_k$ in the target $t$ -- which is then adapted to reflect the target domain statistics (Eq.~(\ref{eq:varUpdates})).

\subsection{Semi-supervised learning (SSL)}

Beyond learning from annotated data in different but related domains, our framework can also be applied in a SSL context to learn from unlabelled data in the same domain to improve performance and/or reduce annotation requirement.  Specifically, images $j$ with known annotations are encoded as described in Sec.~\ref{sec:learning}, while those without annotation are set to $\alpha_{j}^{fg}=0.1~\forall j$, meaning that all topics/classes may potentially occur, but we expect few
simultaneously within one image. Unknown images can include those from
the same pool of classes but without annotation (for which the posterior
$q(\boldsymbol{\theta})$ will pick out the present classes), or those from a completely
disjoint pool of classes (for which $q(\boldsymbol{\theta})$ will
encode only background).


\section{Experiments}
\label{experiments}

\subsection{Datasets, features and settings} 
\label{datasets}

\noindent \textbf{Datasets}\quad We evaluate our model on three datasets, PASCAL VOC \cite{pascalvoc2007}, ImageNet \cite{imagenet_cvpr09} and YouTube-object video \cite{eth_biwi_00905}. The challenging PASCAL VOC 2007 dataset is now widely used for weakly supervised object localisation. A number of variants are used: \textit{VOC07-20} contains all 20 classes from VOC 2007 training set as defined in \cite{Sivaiccv2011} and was used in \cite{Sivaiccv2011,confeccvSivaRX12,zhiyuan12}; \textit{VOC07-6$\times$2} contains 6 classes with Left and Right poses considered as separate giving 12 classes in total and was used in \cite{Deselaers2012,Pandeyiccv2011,Sivaiccv2011,confeccvSivaRX12,zhiyuan12,TangCVPR14}.
The former obviously is more challenging than the latter.  Note that \textit{VOC07-20} is different to the \textit{Pascal07-all} defined in  \cite{Deselaers2012} which actually contains 14 classes and uses the other 6 as fully annotated auxiliary data. We call it \textit{VOC07-14} for consistency, but do not use the other 6 auxiliary classes. 

To evaluate our method in a larger-scale setting, we select all images with bounding box annotation in the ImageNet dataset containing 3624 object categories as in \cite{TangCVPR14}.

We also evaluate our model on videos although it is designed primarily for individual images and does not exploit motion information during learning. Only a simple temporal smoothing post-processing step is introduced (see Sec.~\ref{sub:Object-Localisation}). YouTube-Object dataset \cite{eth_biwi_00905} is a weakly annotated dataset composed of 10 object classes in videos from YouTube.  These 10 classes are a subset of the 20 VOC classes, which facilitate domain transfer experiments. \\

\noindent \textbf{Features}\quad By default, we use only a single appearance feature, namely SIFT to compare directly with most prior WSOL work which uses the same feature. Given an image $j$, we compute $N_{j}$ 128-bin SIFT descriptors, regularly sampled every 5 pixels along both directions, and quantise them into a $2000$-word codebook using K-means clustering. Differently to other bag-of-words (BoW) approaches \cite{LiSocherFeiFei2009,wangbleifeifei08} which then discard spatial information entirely, we then represent
each image $j$ by the list of $N_j$ visual words and corresponding locations
$\{x_{i},l_{ai},l_{bi}\}_{i=1}^{N_j}$ where $\{l_{ai},l_{bi}\}$ are the coordinates of each word. 

We additionally extract two more BoW features at the same regular grid locations to test the feature fusion performance.  They are: (1) Colour-LAB: Colour provides complementary information to SIFT gradients. We quantise colour histograms into three channels (8,16,16) of LAB space and concatenate them to produce a 40 dimensional feature vector. Visual words are then obtained by quantising the feature space using K-means with K=500.
(2) Local binary pattern (LBP) \cite{lbp2002}: 52 bin LBP feature vectors  are computed and quantised into a 500-bin histogram.

\noindent \textbf{Settings and implementation details}\quad For our model, we set the foreground topic number $K^{fg}$ to be equal to the number of classes, and $K^{bg}=20$ for background topics. $\alpha$ is set to 0 or 1 as discussed in Sec.~4. and $\pi^0$ is initialised by Eq.~(8) as described in Sec.~5. $\mu^0$ is initialised with the central of the image area. $\Lambda^0$ is initialised from the half size of the image area. We run Eqs.~(\ref{eq:theta})-(\ref{eq:varUpdates}) for a fixed $100$ VMP iterations. The localisation performance is measured using  CorLoc \cite{eth_biwi_00905,TangCVPR14}: an object is considered to be correctly localised in an given image if the overlap between the localisation box and the ground-truth (any instance of the target class) is greater than 50\%. The CorLoc accuracy is then computed  as the percentage (\%) of correctly localised images for each target class. The same measure has been used in all methods compared in our experiments.


\subsection{Comparison with state-of-the-art}
\label{soa}

\subsubsection{Results on VOC dataset}
\label{sec:cmpOnVOC}

\noindent \textbf{Competitors}\quad We compare our joint modelling approach to the following state-of-the-art competitors:

\vspace{5pt}

\hangafter=1
\setlength{\hangindent}{2em}
\noindent
\textit{Deselaers \etal \cite{Deselaers2012}} A CRF-based multi-instance approach that localises object instances while learning object appearance. They report performance both with a single feature (GIST) and four appearance features (GIST, colour histogram, BoW of SURF, and HOG).

\hangafter=1
\setlength{\hangindent}{2em}
\noindent
\textit{Pandey and Lazebnik \cite{Pandeyiccv2011}} They adapt the fully supervised deformable part-based models to address the weakly supervised localisation problem. 

\hangafter=1
\setlength{\hangindent}{2em}
\noindent
\noindent \textit{Siva and Xiang \cite{Sivaiccv2011}} A greedy search method based on Genetic Algorithm to localise the optimal object bounding box location against a costing function combining the object saliency, intra-class and inter-class cues. 

\hangafter=1
\setlength{\hangindent}{2em}
\noindent \textit{Siva  \etal NM \cite{confeccvSivaRX12}} A simple negative mining (NM) approach which shows that inter-class is a stronger cue than the intra-class one when used properly. 

\hangafter=1
\setlength{\hangindent}{2em}
\noindent \textit{Siva \etal OS \cite{Siva_2013_CVPR}} The negative mining approach above is extended to mine objective saliency (OS) information from a large corpus of unlabelled image. This can be considered as a hybrid of the object saliency approach in \cite{Alexe_TPAMI_2012} and the negative mining work in \cite{confeccvSivaRX12}. 

\hangafter=1
\setlength{\hangindent}{2em}
\noindent \textit{Shi \etal \cite{zhiyuan12}} A ranking based transfer learning approach using an auxiliary dataset to score each candidate bounding box location in an image according to the degree of overlap with the unknown true location.

\hangafter=1
\setlength{\hangindent}{2em}
\noindent \textit{Zhu \etal \cite{zhu2014unsupervised}} An unsupervised saliency guided approach to localise an object in a weakly labelled image in a multiple instance learning framework.

\hangafter=1
\setlength{\hangindent}{2em}
\noindent \textit{Tang \etal \cite{TangCVPR14}} An optimisation-centric approach that uses a convex relaxation of the MIL formulation.

\vspace{0.2cm}
Note that a number of the competitors \cite{Deselaers2012,zhiyuan12,Sivaiccv2011,confeccvSivaRX12,TangCVPR14} used an additional auxiliary dataset that we do not use. Objectness trained on auxiliary data was required by \cite{Deselaers2012,zhiyuan12,Sivaiccv2011,confeccvSivaRX12,TangCVPR14}. Although Shi et al.~\cite{zhiyuan12} evaluated all 20 classes, a randomly selected 10 were used as auxiliary data with bounding-boxes annotation. Pandey and Lazebnik \cite{Pandeyiccv2011} set aspect ratio manually and/or performed cropping on the obtained bounding-boxes.

\begin{table}[ht]
\scriptsize
\begin{center}
\begin{tabular}{l | l | l| l |l | l| l}
\hline
\multicolumn{1}{c|}{\multirow{2}{*}{Method} }&\multicolumn{3}{c|}{Initialisation} & \multicolumn{3}{c}{Refined by detector} \\
\hhline{~------}

&  \textit{6$\times$2} & \  \textit{14} & \ \textit{20} & \textit{6$\times$2}  & \  \textit{14}& \ \textit{20} \\

\hline
\hline
Deselaers \etal  \cite{Deselaers2012} \\
\hhline{~------}
\hspace{10pt}    a. single feature & 35 & 21  & - & 40  & 24  & - \\
\hhline{~------}
\hspace{10pt}    b. all four features & 39 &  22   & -  & 50  &  28   & - \\
\hline
Pandey and Lazebnik \cite{Pandeyiccv2011} $^*$   \\
\hhline{~------}
\hspace{10pt}    a. before cropping & 36.7 & 20.0 & -  & 59.3& 29.0  & - \\
\hhline{~------}
\hspace{10pt}    b. after cropping & 43.7&  23.0 & -  & 61.1 & 30.3 & - \\
\hline
Siva and Xiang \cite{Sivaiccv2011} & 40  & -  & 28.9 & 49  & - & 30.4 \\
\hline
Siva \etal NM \cite{confeccvSivaRX12} & 37.1& -  & 29.0  & 46   & -  & - \\
\hline
Siva \etal OS \cite{Siva_2013_CVPR} & 42.4& -  & 31.1  & 55   & -  & 32.0\\
\hline
Shi \etal \cite{zhiyuan12} $^+$   & 39.7 & -  & 32.1 & - & -  & - \\
\hline
Zhu \etal \cite{zhu2014unsupervised}  & - & -  & - & - & 31   & - \\
\hline
Tang \etal \cite{TangCVPR14} & 39 & - & - & - & - & -\\
\hline
Cinbis \etal \cite{Cinbis_cinbis_2014} &  & - & - & - & - & \textbf{38.8}\\
\hline
\hline
Our-Sampling  &  50.8   & \textbf{32.2} &\textbf{34.1}&  65.5 & \textbf{33.8} &  36.2 \\
\hline
Our-Gaussian  & \textbf{51.5}  & 30.5  &  31.2  &  \textbf{66.1} & 32.5 &  33.4 \\
\hline
\hline
Our-Sampling+prior  & 51.2  & \textbf{33.4}& \textbf{36.1} &  65.9 & \textbf{35.4} & 38.3 \\
\hline
Our-Gaussian+prior  & \textbf{51.8} &31.1 &  33.5  & \textbf{66.7} & 33.0 & 35.8 \\
\hline

\end{tabular}
\end{center}
\caption{Comparison with state-of-the-art competitors on the three variations of the PASCAL VOC 2007 dataset. \footnotesize{$^*$~Requires aspect ratio to be set manually. $^+$~Require 10 out of the 20 classes fully annotated with bounding-boxes and used as auxiliary data.}} 
\label{state-of-art}
\end{table}

\noindent \textbf{Initial localisation } \quad Table \ref{state-of-art} shows that for the initial annotation accuracy our model consistently outperforms all competitors over all three VOC variants, sometimes by big margins. This is mainly due to the unique joint modelling approach taken by our method, and its ability to integrate prior spatial and appearance knowledge in a principled way. Note that the prior knowledge is either based on first principle (spatial and appearance) or  computed from the data without any additional human intervention (appearance). Our two object localisation methods (Our-Sampling and Our-Gaussian) vary in performance over different-sized datasets. Our-Gaussian performs better in the relatively simple datasets (6$\times$2) where most images contain only one object, because our Gaussian location model can compact objects easily in this case. In contrast, Our-Sampling is better in the more complicated situation (20 classes) where many objects co-existing in one image is more common.

\noindent \textbf{Refined by detector}\quad After the initial annotation of the weakly labelled images, a conventional strong object detector can be trained using these annotations as ground truth. The trained detector can then be used to iteratively refine the object location.   We follow \cite{Pandeyiccv2011,Sivaiccv2011} in exploiting a deformable part-based model (DPM) detector\footnote{Version 3.0 is used for fair comparison against most published results obtained using the same version.}  \cite{Felzenszwalb2012partbased} for one iteration to refine the initial annotation. Table \ref{state-of-art} shows that again our model outperforms almost all competitors by a clear margin for all three datasets (see the supplementary material for more detailed per-class comparisons). Very recently, \cite{Cinbis_cinbis_2014} achieved similar performance by training a multi-instance SVM with a more powerful fisher vector based representation. 

\noindent \textbf{With appearance similarity prior}\quad As described before, the proposed framework can exploit the appearance similarity prior across classes. Although the actual appearance similarity between classes is hard to calculate, we can approximate it by computing the relatedness using WordNet semantic tree \cite{wordnetbook1998}. Fig. 5 shows the pairwise relatedness among 20 classes, which is generated using the Lin distance of \cite{Pedersen2004}. The diagonal of the matrix verifies that classes are most similar to themselves. Leaf nodes (blue) correspond to the classes of VOC-20. Classes that inherit from the same subtree should show more similar appearance. A pairwise similarity matrix is then calculated from the tree structure and used to correlate their appearance as explained in Sec.~5. The bottom two rows of Table~\ref{state-of-art} show the localisation accuracy with the appearance similarity prior. It clearly shows that the prior improves the performance of both variants of our model for all experiments.  It is interesting to note that the performance is improved more on VOC-20 than VOC-6$\times$2. This is because there is more opportunity to share related appearance as the number of classes increases. Categories in 6$\times$2 are generally more dissimilar, so there is less benefit to the correlation.

\begin{figure}[t]
\begin{minipage}[b]{4.3cm}
\begin{subtable}{1\textwidth}
\includegraphics[height=3.4cm]{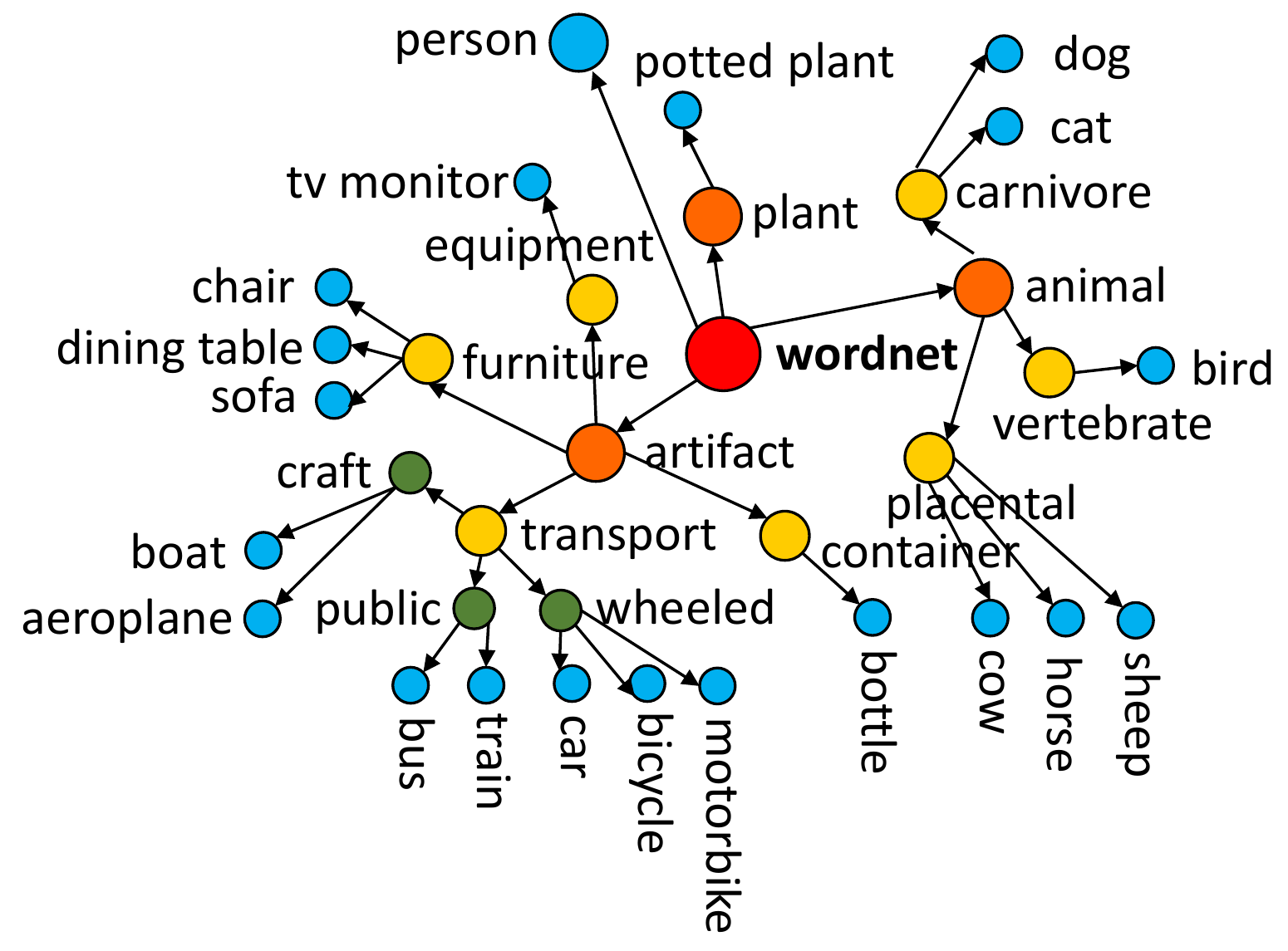}
\caption{}
\end{subtable}
\end{minipage}
\hspace{0.1cm}
\begin{minipage}[b]{3.7cm}
\begin{subtable}{1\textwidth}
\includegraphics[height=3.7cm]{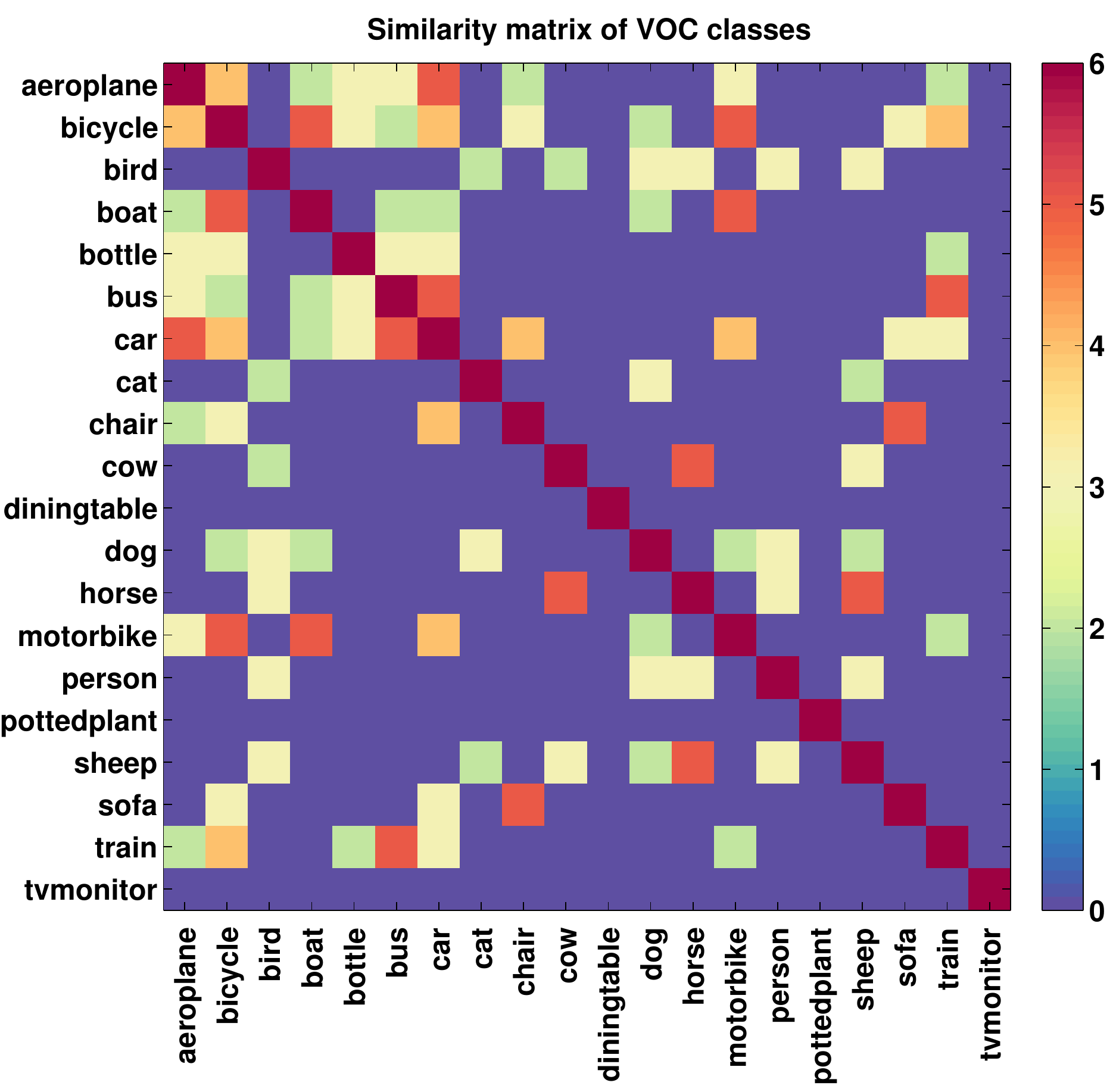}
\caption{}
\end{subtable}
\end{minipage}
\caption{ (a) A  hierarchical structure of the 20 PASCAL VOC classes using WordNet. (b) The class similarity matrix. }
\label{fig:knowledge}
\end{figure}

\begin{figure*}[ht!]
\begin{center}
   \includegraphics[width=\linewidth]{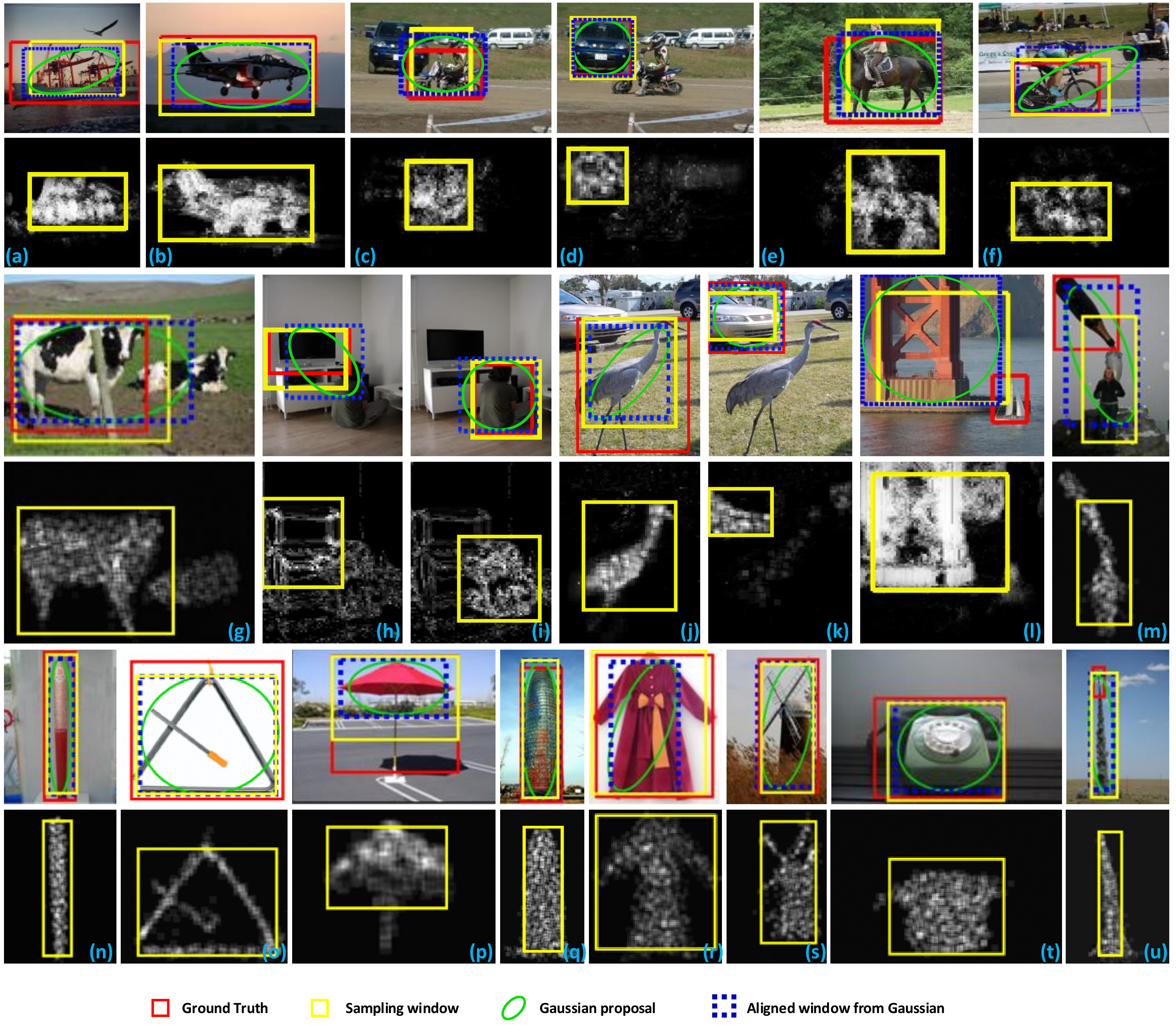}
\end{center}
\vspace{-0.3cm}
 \caption{Top row in each subfigure: examples of object localisation using our-sampling and our-Gaussian. Bottom row: illustration of what is learned by the object (foreground) topics via heat map (brighter means object is more likely). The first four rows show some examples of PASCAL VOC and last two rows are selected from ImageNet.  }
\label{fig:objecttopic}
\end{figure*}

\noindent \textbf{What has been learned}\quad Fig.~\ref{fig:objecttopic} gives examples of the localisation results and illustrates what has been learned for the foreground object classes. For the latter, we show the response of each learned object topic (i.e.~the posterior probability of the topic given the visual word) as a gray-level image, or heat map (the brighter, the higher probability that the object is present at each image location).  These examples show that the foreground topics indeed capture what each object class looks like and can distinguish it from the background and between different object classes. For instance,  Fig.~\ref{fig:objecttopic}{(c)} shows that the motorbike heat map is quite accurately selective, with minimal response obtained on the other vehicular clutter. Fig.~\ref{fig:objecttopic}{(e)} indicates how the Gaussian can sometimes give a better bounding box. The opposite is observed in Fig.~\ref{fig:objecttopic}{(f)} where the single Gaussian assumption is not ideal when the foreground topic has less a compact response. Selectivity is illustrated by Fig.~\ref{fig:objecttopic}{(c,d)}, Fig.~\ref{fig:objecttopic}{(h,i)} and Fig.~\ref{fig:objecttopic}{(g,k)}, which show the same images, but with detection results for different co-occurring objects. In each case, the relevant object has been successfully selected while ``explaining away'' the potentially distracting alternative. Our method may fail if the background clutter or objects of no interest dominates the image (Fig.~4(l,m,u)). For example, in Fig.~\ref{fig:objecttopic}{(l)}, a bridge structure resembles the boat in Fig.~\ref{fig:objecttopic}{(a)} resulting strong response from the boat topic, whilst the actual boat, although picked up, is small and overwhelmed by the false response.

\begin{figure}[t]
\begin{center}
   \includegraphics[width=\linewidth]{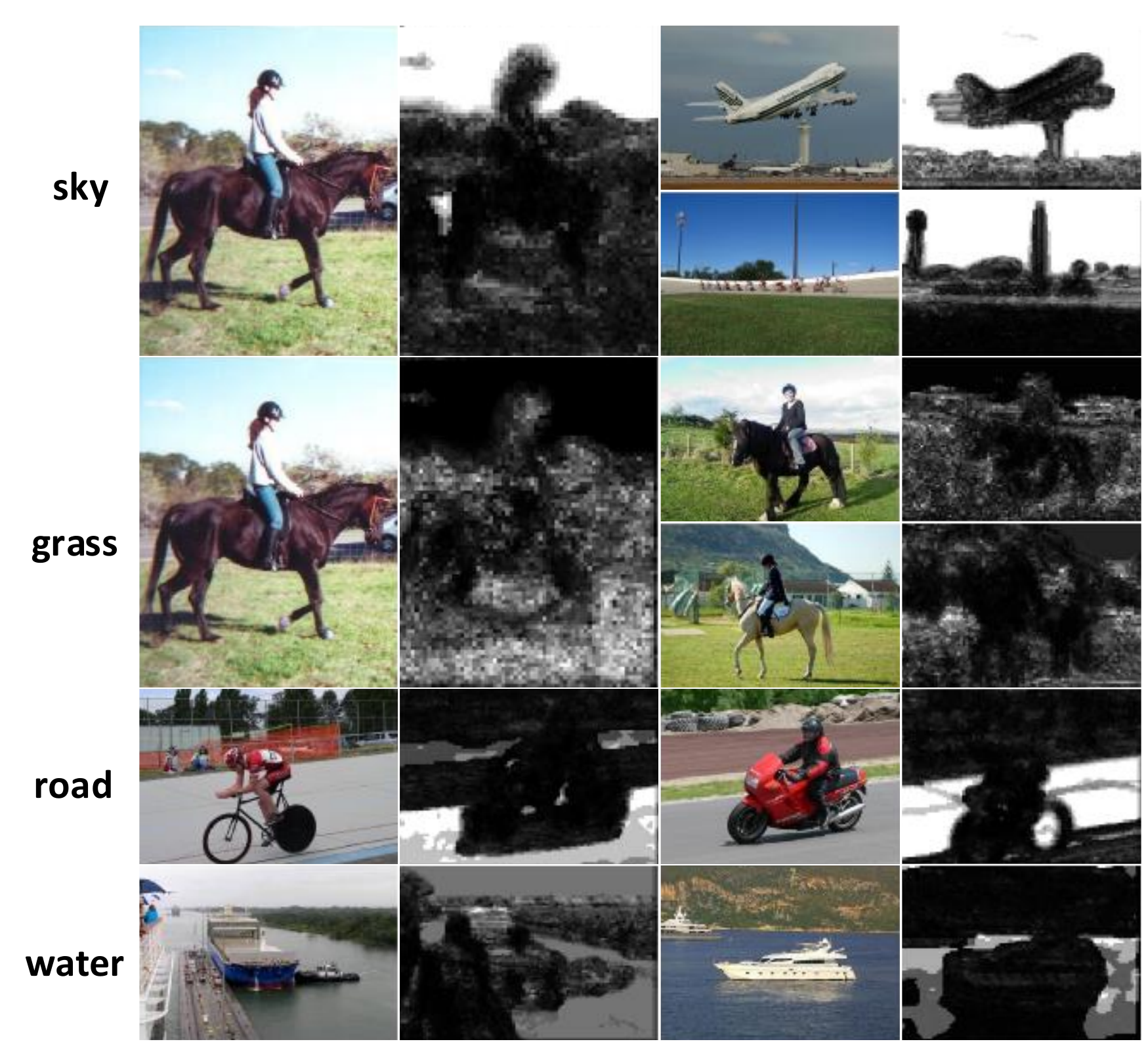}
\end{center}
   \caption{Illustration of the learned background topics.} 
   
\label{fig:backgroundtopic}
\end{figure}

A key strength of our framework is explicit modelling of background without any supervision. This allows background pixels to be explained, reducing confusion with foreground objects and hence improving localisation accuracy. This is illustrated in Fig.~\ref{fig:backgroundtopic} via plots of the background topic response (heat map). It illustrates qualitatively that some background topics 
are often correlated with common semantic background components such as sky, grass, road and water, despite none of these being annotated. 

\noindent \textbf{Weakly supervised detector}\quad The ultimate goal of weakly supervised object localisation is to learn a weakly supervised detector. This is achieved by feeding the localised objects into an off-the-shelf detector training model. The deformable part based  model (DPM) in \cite{Felzenszwalb2012partbased} is used and this weakly supervised (WS) detector is compared against a fully supervised (FS) one with the same DPM model (version 3.0). 
Specifically, Table \ref{tab:detection} compares the mean average precision (mAP) of detection performance on both VOC-6$\times$2 and VOC-20 test datasets among previous reported WS detector results, ours and the fully supervised detector \cite{Felzenszwalb2012partbased}. Due to the better localisation performance on the weakly supervised training images, our approach is able to reduce the gap between the WS detector and the FS detector. The detailed per-class result is included in the supplementary material and it shows that for classes  with high localisation accuracy (e.g.~bicycle, car, motorbike, train), the WS detector is often as good as the the FS one, whilst for those with very low localisation accuracy (e.g.~bottle and pottedplant), the WS detector fails completely. 

\begin{table}[ht]
\footnotesize
\begin{center}
\setlength{\tabcolsep}{0.2em}
\begin{tabular}{l || l | l |l || l ||l}

\hline
Method &  Deselaers \cite{Deselaers2012} & Pandey \cite{Pandeyiccv2011} & Siva \cite{Sivaiccv2011} & \textbf{Ours} & Fully Supervised \\

\hline
\hline
6$\times$2  & 21   &   20.8 & - & 26.1 &33.0\\
\hline
20  &  -  &   -  &13.9 & 17.2 &26.3\\
\hline

\end{tabular}
\end{center}
\caption{Performance  of strong detectors trained using annotations obtained by different WSOL methods}
\label{tab:detection}
\end{table}

\subsubsection{Results on ImageNet dataset}

\begin{table}[ht]
\footnotesize
\begin{center}
\setlength{\tabcolsep}{0.2em}
\begin{tabular}{l || l}

\hline
Method &  Initialisation \\
\hline
\hline
Alexe \etal \cite{Alexe_TPAMI_2012}  & 37.4\\
\hline
Tang \etal \cite{TangCVPR14} & 53.2\\
\hline
Our-Sampling & \textbf{57.6}\\
\hline

\end{tabular}
\end{center}
\caption{Initial annotation accuracy on ImageNet dataset}
\label{tab:imagenet}
\end{table}

Table \ref{tab:imagenet} shows the initial annotation accuracy of different methods for the much larger $3624$-class ImageNet dataset. Note that the result of Alexe \etal \cite{Alexe_TPAMI_2012} is taken from the Table 4 in \cite{TangCVPR14}.  Although the annotation accuracy could be further improved by training an object detector to refine the annotation as shown in Table 2, this step is omitted in our experiment as none of the competitors attempted it.
For such a large scale learning problem, loading all the image features into the memory is a challenge for our joint learning method. A standard solution is taken, that is, to process in batches of 100 classes. Joint learning is performed within each batch  but not across batches; our model is thus not used to its full potential. Table \ref{tab:imagenet} shows that our method achieves the best result (57.6\%). Note that \cite{Alexe_TPAMI_2012} is a very simple baseline as it simply takes the top-scoring objectness box. Recently more sophisticated transfer-based techniques \cite{Guillaumin_cvpr12} and \cite{Vezhnevets_CVPR_2014} were evaluated on ImageNet. But their results were obtained on a different subset of ImageNet, thus not directly comparable here.

To investigate the effect of the similarity prior in this larger dataset, we randomly choose 500 small (containing around 100 images each) leaf-node classes from ImageNet for joint-learning with an inter-class similarity prior. This was the largest dataset size that could simultaneously fit in the memory of our platform\footnote{Our learning algorithm could potentially be modified to process all $3624$ classes in batches.}. Performing joint learning with inter-class correlation on this ImageNet subset, we achieve 58.8\% annotation accuracy on the 500 classes compared to 55.4\% without using the similarity prior.

\subsubsection{Results on YouTube-object dataset}
\label{sec:cmpOnVideo}

Our main competitors on YouTube-Object (YTO) are \cite{eth_biwi_00905} and \cite{TangECCV14}. Prest \etal \cite{eth_biwi_00905} first performed spatio-temporal segmentation of video into a set of 3D tubes, and subsequently searched for the best object location. Very recently, \cite{TangECCV14} simultaneously localised objects of the same class across a set of video clips (co-localisation) with the Frank-Wolfe Algorithm.  Note that there are some recently published studies on weakly supervised object segmentation from video \cite{Tang_CVPR2013}. This is not directly comparable as they did not report results based on the standard YTO bounding-box annotations. Two variants of our model are compared here: Our-sampling is the method evaluated above for individual images. Used here, it ignores the temporal continuity of the video frames in a video. Our-smooth is the simple extension of our sampling for video object localisation. As described in Sec.~\ref{sub:Object-Localisation}, temporal information is used to  enforce a smooth change of object location over consecutive frames. The way  temporal information is exploited is thus much less elaborative than that in \cite{eth_biwi_00905}. For all methods compared, We evaluate  localisation performance on the key frames which are provided with ground truth labels by \cite{eth_biwi_00905}. 

\begin{figure}[t]
 \centering
  \includegraphics[width=\linewidth]{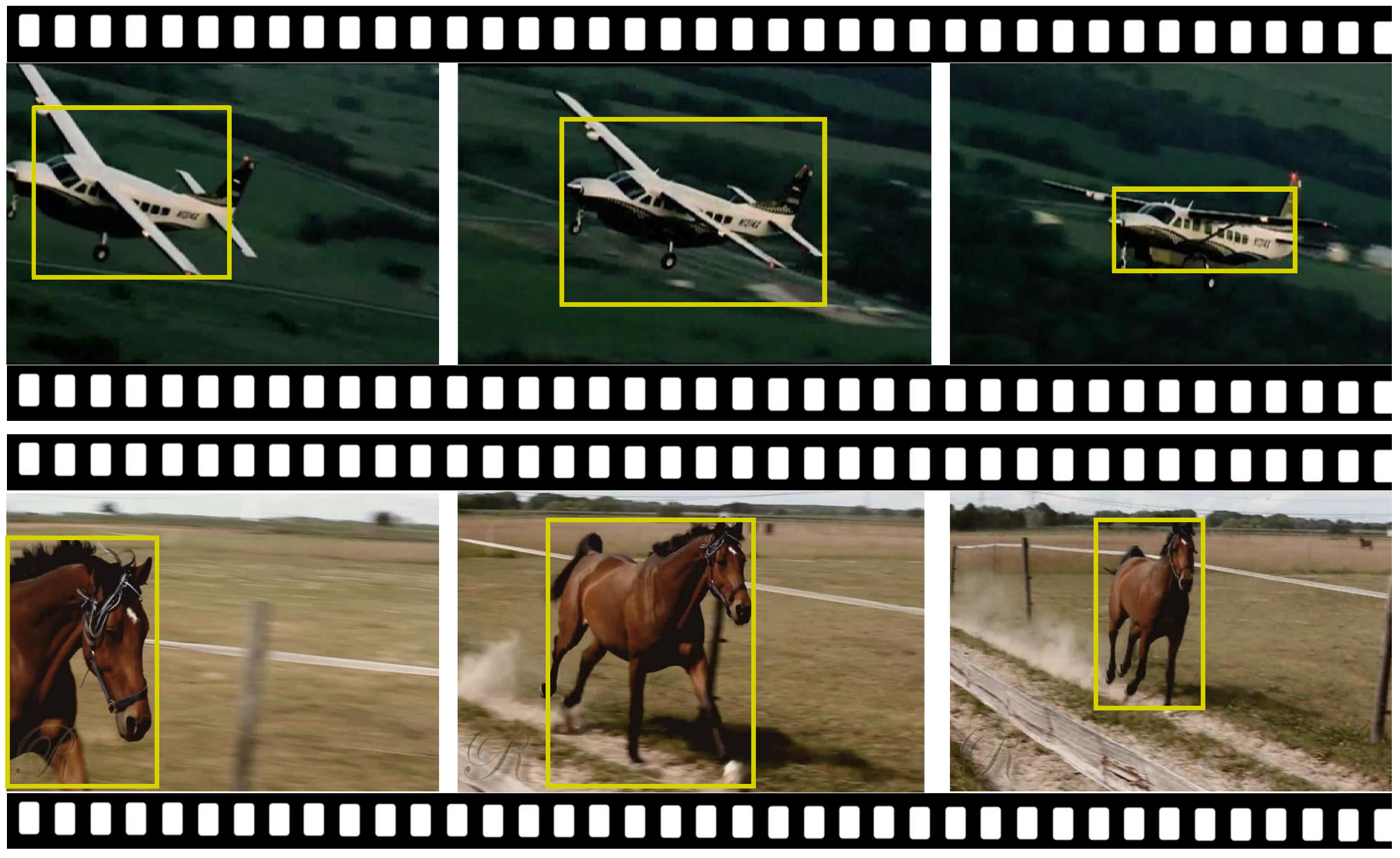}
   \caption{Examples of video object localisation}
  \label{ytb_result}
\end{figure}

Table~\ref{tab:youtube} shows that even without using any temporal information and operating on  key frames only, Our-sampling outperforms the method in \cite{eth_biwi_00905}. Our-Smooth  further improves the performance and the localisation accuracy of 32.2\% is very  close to the upper bound result (34.8\%) suggested by \cite{eth_biwi_00905}, which is the best possible result from oracle tube extraction. Fig.~\ref{ytb_result} shows some examples of video object localisation using Our-Smooth. We note that all these results have been exceeded (50.1\% accuracy) recently by a model purposefully designed for video segmentation \cite{Papazoglou_iccv13}, which performed much more intensive spatio-temporal modelling and used superpixel segmentation within each frame and motion segmentation across frames.

\begin{table}[h]
\footnotesize
\centering
\begin{tabular}{l| l| l |c|c|c}
\hline
Categories & \cite{eth_biwi_00905} & \cite{TangECCV14} & Our-Sampling & Our-Smooth & \cite{Papazoglou_iccv13} \\
\hline
\hline
aeroplane & 51.7& 27.5&40.6 &45.9 & 65.4 \\
\hline
bird & 17.5 & 33.3&39.8 & 40.6 & 67.3 \\
\hline
boat & 34.4 &27.8& 33.3& 36.4 & 38.9 \\
\hline
car & 34.7 & 34.1 & 34.1&33.9 & 65.2\\
\hline
cat & 22.3 & 42.0&35.3 &35.3   & 46.3 \\
\hline
cow & 17.9  & 28.4&18.9 &22.1  & 40.2 \\
\hline
dog & 13.5  &35.7& 27.0 &27.2 & 65.3 \\
\hline
horse &26.7 &35.6 & 21.9 &25.2 & 48.4 \\
\hline
motorbike & 41.2 & 22.0 & 17.6 &20.0 & 39.0\\
\hline
train & 25.0 & 25.0&  32.6 &35.8 & 25.0 \\
\hline
\hline
Average & 28.5 & 31.1& 30.1 & 32.2 & 50.1\\
\hline
\end{tabular}
\caption{Performance comparison on YouTube-object}
\label{tab:youtube}
\end{table}

\subsection{Bayesian domain adaptation}
\label{sec_dt}

We next evaluate the potential of our model for weakly supervised cross-domain transfer learning using the YouTube-Object and VOC07-10 as the two domains (we choose the same 10 classes from the VOC07-20 as in YouTube-Object). One domain contains continuous and highly varying video data, and the other contains high resolution but cluttered still images.  We consider following two non-transfer baselines: 

\vspace{0.5em}

\hangafter=1
\setlength{\hangindent}{2em}
\noindent \textit{YTO, VOC} The first baseline is the original performance on YouTube-Object and VOC07-10 classes, solely using target domain data. \textit{YTO} is exactly the same as Our-Sampling  described in Sec.~\ref{sec:cmpOnVideo}, while \textit{VOC} is trained with 10 classes from VOC07-20 using the same setting described in Sec.~\ref{sec:cmpOnVOC}.

\hangafter=1
\setlength{\hangindent}{2em}
\noindent \textit{All$\rightarrow$YTO, All$\rightarrow$VOC} The second baseline  simply combines the training data of YouTube-Object and VOC. One model trained with these two domains' data is used to localise object on YouTube-Object (\textit{A$\rightarrow$Y}) and VOC07-10 (\textit{A$\rightarrow$V}).

\vspace{0.5em}

We consider two directions of  knowledge transfer between YouTube-Object and VOC07-10, and compare the above baselines with our domain adaptation method: \textit{V$\rightarrow$Y} is initialised with an appearance  prior transferred from VOC07-10, and adapted on the YTO data. On the contrary, \textit{Y$\rightarrow$V} adapts the YTO appearance prior to VOC07-10. Table~\ref{tab:youtube_transfer} shows that our Bayesian domain adaption method  performs better than the baselines on both YouTube-Object and VOC07-10. In contrast, the standard combination (A$\rightarrow$Y and A$\rightarrow$V) shows little advantage over solely using target domain data. Note that unlike prior studies of video$\rightarrow$image \cite{eth_biwi_00905} or image$\rightarrow$video \cite{Tang_NIPS2012} that adapt detectors with fully labelled data, our task is to adapt weakly labelled data. 

\begin{figure}[h]
\begin{center}
   \includegraphics[width=0.6\linewidth]{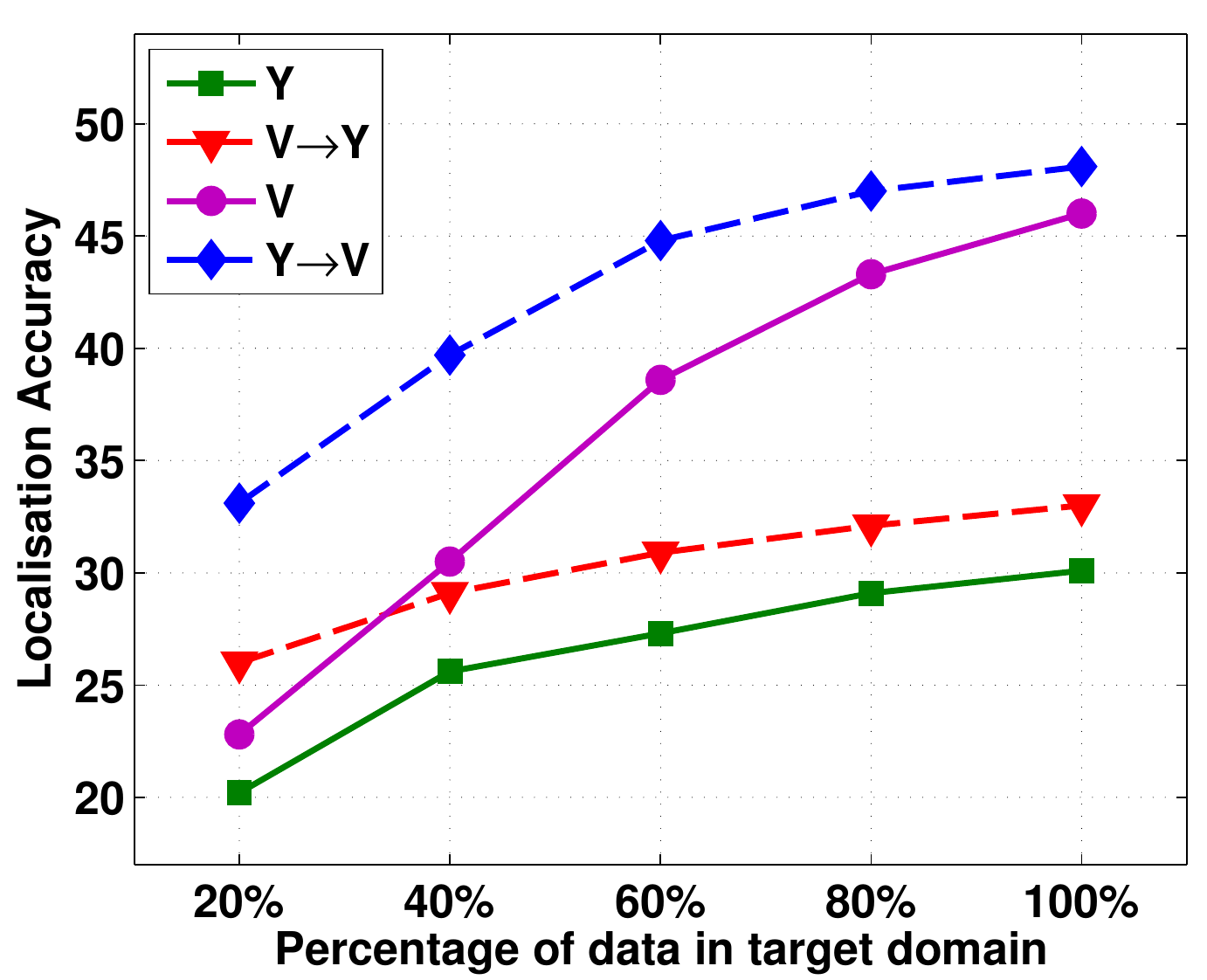}
\end{center}
\vspace{-0.3cm}
   \caption{Domain adaptation provides more benefit with fewer target domain samples.}
\label{fig:domaintransfer}
\end{figure}

We also vary the amount of target domain data and evaluate its effect on the domain transfer performance. Fig.~\ref{fig:domaintransfer} shows that our model provides a bigger margin of benefit given less target domain data. This can be easily understood because with a small quantity of training examples there is insufficient data to learn the object appearance well and the impact of the knowledge transfer is thus more significant. 

\begin{table}[t]
\setlength{\tabcolsep}{0.5em}
\footnotesize
\centering
\begin{tabular}{l||l| l| l|| l|l|l}

\hline

\multirow{2}{*}{Categories } &  \multicolumn{3}{c||}{YTO}  &\multicolumn{3}{c}{VOC} \\
\cline{2-7}
& \hspace{0.05cm} {Y} & {A$\rightarrow$Y} &  {V$\rightarrow$Y}  &\hspace{0.05cm} {V} &  {A$\rightarrow$V}  &{Y$\rightarrow$V} \\
\hline
\hline
aeroplane & 40.6  &40.8& \textbf{45.8} & 57.5& 58.1& \textbf{58.7} \\ 
\hline
bird & 39.8  &\textbf{40.3}& 38.8 & 29.8 &30.5& \textbf{33.7} \\
\hline
boat&     33.3  &33.4& \textbf{38.8} & 28.0 &27.9& \textbf{29.0} \\ 
\hline
car & \textbf{34.1}  &33.9 &33.6 & 39.1 &39.1& \textbf{44.4} \\ 
\hline
cat &  35.3  &35.3& \textbf{38.8} & 59.0 &\textbf{59.3}& 58.6 \\
\hline
cow & 18.9  &19.0& \textbf{27.7} & 36.7 & 36.9&\textbf{38.9} \\
\hline
dog & 27.0  &\textbf{27.1}& 26.7 & 46.5 &47.4& \textbf{48.3} \\ 
\hline
horse &21.9  &22.1& \textbf{26.1} & 53.2 &53.5 &\textbf{55.5} \\
\hline
motorbike & 17.6  &\textbf{17.9}& 17.5 & 55.6 & 55.2&\textbf{58.1} \\
\hline
train &  32.6  & 32.6&\textbf{36.2} & 54.7 &54.5& \textbf{56.3} \\
\hline
\hline
\textbf{Average}  &30.1 &30.2& \textbf{33.0} & 46.0 & 46.2&\textbf{48.1} \\ 
\hline
\end{tabular}
\caption{Cross-domain transfer learning results}
\label{tab:youtube_transfer}
\end{table}

\subsection{Semi-supervised Learning}

\label{semi-supe}

One important advantage of our model is the ability to utilise unlabelled data to further reduce the manual annotation requirements. To demonstrate this we randomly select $10\%$ of the \textit{VOC07-6$\times$2} data as our weakly labelled training data, and then vary the additional unlabelled data used. Note that 10\% labelled data corresponds to around only \textit{5 weakly labelled images per class} for the  \textit{VOC07-6$\times$2} dataset, which is significantly less than what any previous method has exploited. Two evaluation procedures are considered: (i) Evaluating localisation performance on the initially annotated $10\%$ (standard WSOL task); and (ii) WSOL performance on the held out  \textit{VOC07-6$\times$2} test set\footnote{To localise objects in a test image, we only need to iterate Eqs.~(\ref{eq:theta})-(\ref{eq:y}) instead of (\ref{eq:theta})-(\ref{eq:varUpdates}). That is, the object appearance is considered fixed and does not need to be updated. This both reduces the cost of each iteration and also makes convergence more rapid.}. 
The latter corresponds to an online application scenario where the localisation model is trained on one database and needs to be applied online to localise objects in incoming weakly labelled images. We vary the additional data across a combination of four conditions: (1) $6R$: add the remaining 90\% of data for the 6 target classes but without labels, (2) $100U$: add all images from 100 unrelated ImageNet classes without labels, (3) $6R+100U$: add both of the above. There are two questions to answer: Whether the model can exploit the related data when it comes without labels (6R), and whether it can avoid being confused by a vast quantity of unrelated data (100U).

\begin{figure}[t]
\begin{center}
   \includegraphics[width=\linewidth]{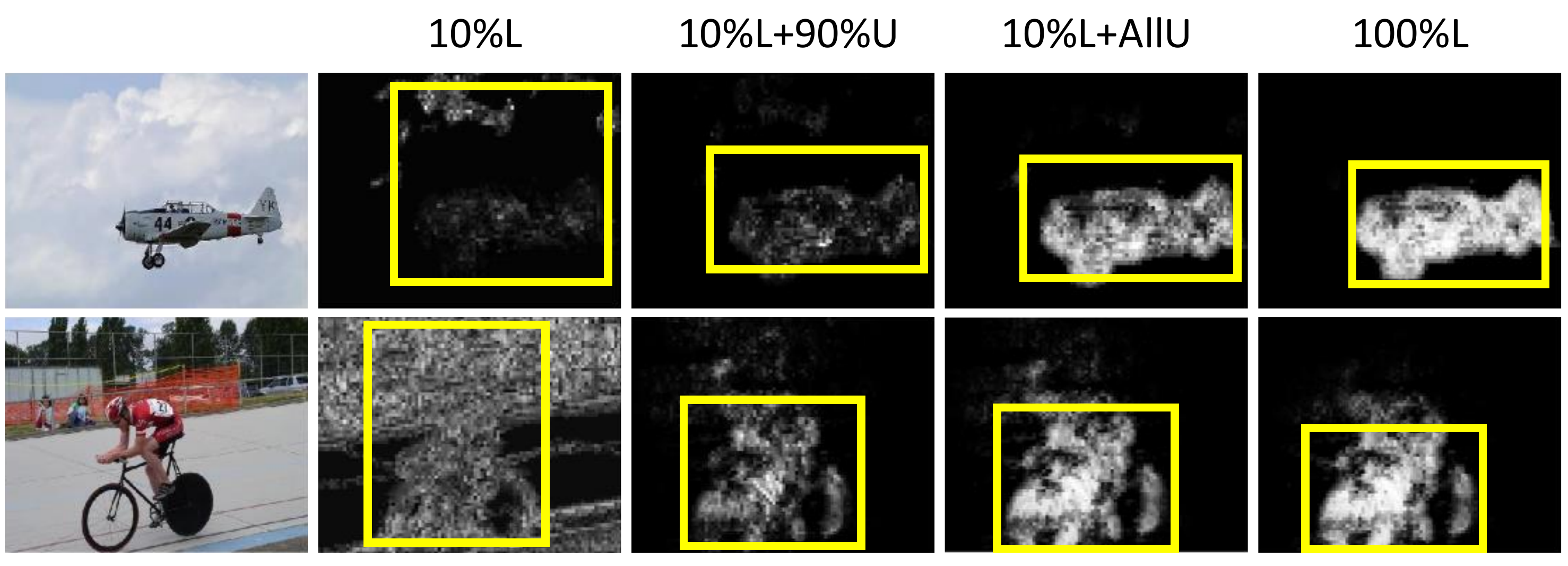}
\end{center}
\vspace{-0.3cm}
   \caption{Unlabelled data improves foreground heat maps.}
\label{fig:semilearned}
\end{figure}

The results are shown in Table~7, where the ratio of relevant to irrelevant data in the additional unlabelled samples is shown in the second column. From the results, we can draw the following conclusions: (1) As expected, the model performs poorly with little data (10\%L). However it improves significantly with some relevant but unlabelled data (the standard SSL setting, 10\%L+6R). Moreover, this SSL result is almost as good as when all the data is labelled (100\%L). (2) If \emph{only} irrelevant data is added to the small labelled seed, not only does the performance not degrade, but it increases noticeably (10\%L vs. 10\%L+100U). (3) If both relevant and irrelevant data are added -- corresponding to the realistic scenario where an automatic process gathers a pool of potentially relevant data which, without any screening, will be a mix of relevant and irrelevant data to the target problem. In this case the performance improves to not far off the fully annotated case (10\%L vs. 10\%L+6R+100U vs. 100\%L). As expected, the performance of 10\%L+6R+100U is  weaker than 10\%L+6R -- if one manually goes through the unlabelled data and removes the irrelevant ones and leave only the relevant ones, it would certainly benefit the model. But it is noted that the decrease in performance is small (47.1\% to 43.5\%).  (4) If the irrelevant data is added to the fully annotated dataset, the  performance improves slightly  (100\%L vs. 100\%L+100U), which shows that our model is robust to this potential distraction from the large amount of unlabelled and irrelevant data. This is expected in SSL, which typically benefits only when the amount of labelled data is small. These results show that our approach has good promise for effective use in  realistic scenarios of learning from only few weak annotations and a large volume of only partially relevant unlabelled data. This is illustrated visually in Fig.~\ref{fig:semilearned}, where unlabelled data helps to learn a better object model. Finally, the similarly good results on the held-out test set verify that our model is indeed learning a good generalisable localisation mechanism and is not over-fitted to the training data. 


\begin{table}[ht]
\footnotesize
\begin{center}
\begin{tabular}{l|l || c | c  }
\hline
\multicolumn{2}{c}{VOC07-$6\times2$} & \multicolumn{2}{c}{Data for Localisation}\\
\hline
Data for Training & ratio of R:U &  \textit{10\%L} & \textit{Test set} \\

\hline
\hline
\textit{10\%L}  & -& 27.1   &   28.0  \\
\hline
\textit{10\%L+6R} & 1 & 47.1  &   42.3   \\
\hline
\textit{10\%L+100U} & 0&  35.8 &  32.4    \\
\hline
\textit{10\%L+6R+100U} & 0.04& 43.5   & 38.1  \\
\hline
\hline
\textit{100\%L} & -& 50.3   &  46.2  \\
\hline
\textit{100\%L+100U} & 0&  50.7 & 47.5\\
\hline
\end{tabular}
\end{center}
\caption{Localisation performance of semi-supervised learning using \textit{Our-Sampling}}
\label{tab:sslBar}
\end{table}

\subsection{Computational cost}
\label{cost}
Our model is efficient both in learning and inference, with a complexity $\mathcal{O}(NMK)$ for $N$ images, $M$ observations (visual words) per image, and $K$ classes. The experiments were done on a 2.6GHz PC with a single-threaded Matlab implementation. Training the model on all 5,011 VOC07 images required 3 hours and a peak of 6 GB of memory to learn a joint model for 20 classes.
Our Bayesian topic inference process not only enables prior knowledge to be used, but also achieves 10-fold improvements in convergence time compared to EM inference used by most conventional topic models with point-estimated Dirichlet topics. Online inference of a new test image took about 0.5 seconds. After model learning, for object localisation in training images, direct Gaussian localisation is effectively free and  heat-map sampling took around 0.6 seconds per image. These statistics compare favourably to alternatives: \cite{Deselaers2012}  reported 2 hours to train 100 images; while our Matlab implementations of \cite{confeccvSivaRX12},  \cite{Sivaiccv2011} and \cite{blei2003annotated_model} took 10, 15 and 20 hours respectively to localise objects for all 5,011 images.


\section{Conclusion and Future work} We have presented an effective and efficient model for weakly-supervised object localisation (WSOL). Our approach surpasses the performance of  prior methods and obtains state-of-the-art results on PASCAL VOC 2007 and ImageNet datasets. It can also be applied to the YouTube-Object dataset, and to domain transfer between these image and video datasets.
With joint multi-label modelling, instead of independent learning in previous work, our model enables: (1) exploiting multiple object co-existence within images, (2) learning a single background shared across classes and (3) dealing with large scale data more efficiently than prior approaches. Our generative Bayesian formulation, enables a number of novel features: (1) integrating appearance and geometry priors, (2) exploiting inter-category appearance similarity and (3) exploiting different but related datasets via domain adaptation. Furthermore, it is able to use (potentially easier to obtain) unlabelled data with a challenging mix of relevant and irrelevant images to obtain an reasonable localiser when labelled data are in short supply for the target classes.

In this study we showed the usefulness of top-down, cross-class and domain transfer priors -- demonstrating the model's potential to scale learning through transfer \cite{zhiyuan12,Guillaumin_cvpr12,Kuettel2012}. These contributions bring us significantly closer to the goal of scalable learning of strong models from weakly-annotated non-purpose collected data on the Internet. 

It is  worth pointing out that apart from adding a few new features (e.g.~foreground-background topic separation and effective supervision via topic clamping), our generative Bayesian topic model is not fundamentally different from existing topic models used for image understanding \cite{LiSocherFeiFei2009,Philbinijcv2010}. Nevertheless, state-of-the-art WSOL performance is obtained compared with  more popular, more highly engineered and complex, and slower discriminative models. This not only shows the importance of the change of paradigm from independent discriminative learning to joint generative learning, but also suggests that sometimes it is not necessary to invent a completely new model; finding the missing ingredients that make an existing model work can be equally important. 

Possible directions for future work include: automatically determining the optimal number of topics $K$ \cite{sudderth2008tdp_visual},  learning a deeper multi-layered \cite{sudderth2008tdp_visual} model by exploiting parts \cite{Crandalleccv06,sudderth2008tdp_visual} and attributes \cite{fu2012attribsocial} rather than the current flat model; learning rather than pre-defining object-appearance similarity \cite{Salakhutdinov2011cvpr}; and learning from realistically noisy non-purpose collected labels \cite{fu2012attribsocial}.



%

\bibliographystyle{IEEEtran}
\bibliography{egbib_zhiyuan}

\vspace{-0.1cm}
\begin{biography}[{\includegraphics[width=1in,height=1.25in,clip,keepaspectratio]{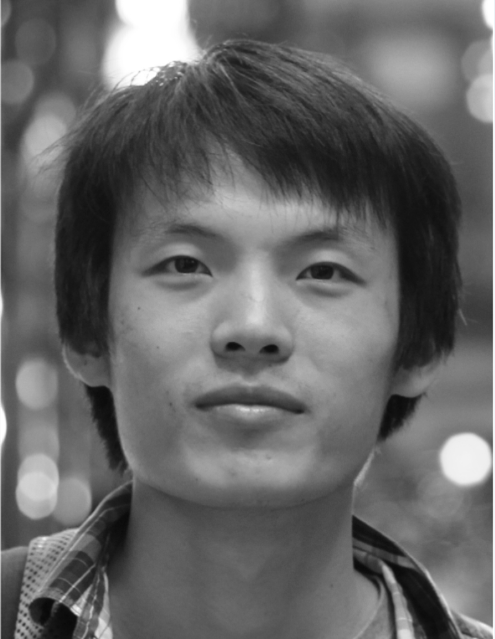}}]{Zhiyuan Shi} received the BEng degree in electronic engineering and computer science from Beijing University of Posts and Telecommunications in 2011. He is currently a PhD student in the School of Electronic Engineering and Computer Science, Queen Mary University of London. His research interests include weakly supervised learning, topic model, object localisation and attribute learning.
\end{biography}

\begin{biography}[{\includegraphics[width=1in,height=1.25in,clip,keepaspectratio]{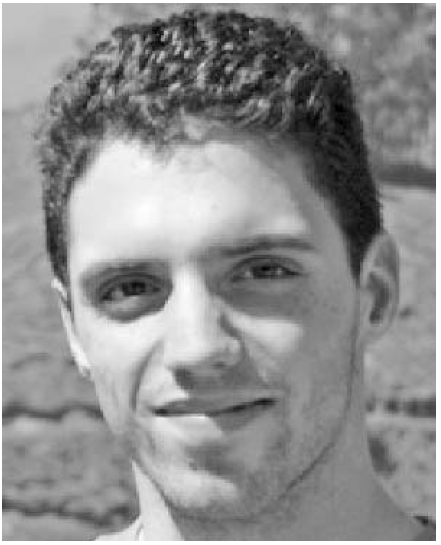}}]{Timothy M. Hospedales} received  the  PhD degree in neuroinformatics from the University of Edinburgh in 2008. He is currently a lecturer (assistant professor) of computer science at Queen Mary University of London. His research interests include probabilistic modelling and machine learning applied variously to problems in computer vision, data mining, interactive learning, and neuroscience. He has published more than 20  papers  in  major  international journals and conferences. He is a member of the IEEE.
\end{biography}

\begin{biography}[{\includegraphics[width=1in,height=1.25in,clip,keepaspectratio]{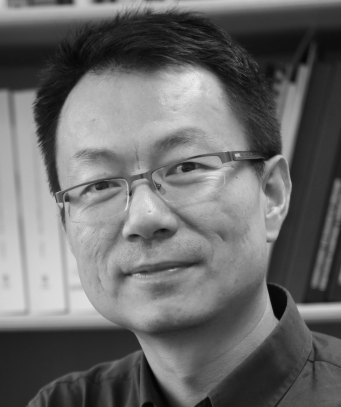}}]{Tao Xiang} received the PhD degree in electrical and computer engineering from the National University of Singapore in 2002. He is currently a reader (associate professor) in the School of Electronic Engineering and Computer Science, Queen Mary University of London. His research interests include computer vision,  machine learning, and data mining. He has published over 100 papers in international journals and conferences and co-authored a book, Visual Analysis of Behaviour: From Pixels to Semantics.
\end{biography}

\end{document}